\documentclass{article} %
\usepackage[final]{Template-2026/colm2026_conference}
\usepackage[T1]{fontenc}

\usepackage{microtype}
\usepackage{hyperref}
\usepackage{url}
\usepackage{booktabs}

\usepackage{lineno}
\usepackage{etoc}
\usepackage{twemojis}
\usepackage{pgfplots,pgfplotstable}
\usepackage[most,skins,theorems]{tcolorbox}
\tcbuselibrary{listings}
\usetikzlibrary{patterns}
\usepackage{algorithm}
\usepackage{algpseudocode}
\usepackage{tabularx}
\usepackage[normalem]{ulem}
\usepackage{soul}
\usepackage{amsmath,amssymb,amsfonts}
\usepackage{cleveref}
\usepackage{enumitem}
\usepackage{wrapfig}

\usepgfplotslibrary{groupplots}
\definecolor{seabornBlue}{HTML}{3E6487}
\definecolor{seabornGreen}{HTML}{45A77B}

\pgfplotsset{compat=1.18}

\tcbset{
  aibox/.style={
    width=\linewidth,
    top=6pt,
    bottom=0pt,
    colback=blue!6!white,
    colframe=black,
    colbacktitle=black,
    enhanced,
    center,
    attach boxed title to top left={yshift=-0.1in,xshift=0.15in},
    boxed title style={boxrule=0pt,colframe=white,},
  }
}
\newtcolorbox{AIbox}[2][]{aibox,arc=5pt,title=#2,#1}
\newtcolorbox{examplebox}[1][]{
    colback=white,
    arc=0pt,
    boxrule=0pt,
    boxsep=2pt,
    left=4pt,
    right=4pt,
    top=4pt,
    bottom=4pt,
    #1 %
}

\definecolor{darkblue}{rgb}{0, 0, 0.5}
\hypersetup{colorlinks=true, citecolor=darkblue, linkcolor=darkblue, urlcolor=darkblue}

\newcommand{\benchmark}[0]{\textsc{MT-PingEval}}

\title{\texttwemoji{ping pong}\benchmark: Evaluating Multi-Turn Collaboration with Private Information Games}

\author{Jacob Eisenstein \; Fantine Huot \; Adam Fisch \; Jonathan Berant \; Mirella Lapata \\
Google DeepMind\\
\texttt{\{jeisenstein,fantinehuot,fisch,joberant,lapata\}@google.com}}

\begin{document}

\ifcolmsubmission
\linenumbers
\fi

\maketitle
\lhead{Published as a conference paper at COLM 2026}

\begin{abstract}
We present a scalable and verifiable methodology for evaluating language models in multi-turn interactions, using a suite of collaborative games that require effective communication about private information. This enables an interactive scaling analysis, in which a fixed token budget is divided over a variable number of turns. We find that language models often fail to use interactive collaboration to improve over the non-interactive baseline in which one agent summarizes its information and the other agent immediately acts, despite substantial headroom. This suggests that state-of-the-art models still suffer from significant weaknesses in planning and executing multi-turn collaborative conversations. We analyze the linguistic features of these dialogues, assessing the roles of sycophancy, information density, and discourse coherence. While there is no single linguistic explanation for the collaborative weaknesses of contemporary language models, we note that humans achieve comparable task success at superior token efficiency by producing more coherent dialogues. The proactive management of private information is a defining feature of real-world communication, and this work is designed to drive further progress on this capability. 
\end{abstract}

\section{Introduction}
Multi-turn conversation is a core capability for language models, but it is difficult to measure. The primary challenge lies in creating realistic conditions in which (1) dialogue is necessary, yet (2) performance is verifiable. Many existing evaluations focus  on scripted scenarios where a human participant (or more often, a user simulator that is backed by another language model) is given a predefined goal or set of goals, and is asked to play out a multi-turn interaction with an AI assistant~\citep[e.g.,][]{shao2024collaborative,laban2025llms}. There are at least two major drawbacks to this style of evaluation. First, it is fundamentally asymmetric:  the assistant generates a series of proposals (e.g., a travel itinerary, some software code) that the human critiques until reaching a satisfactory output. In contrast, human-human interactions usually require \emph{both} participants to proactively shape the direction of the dialogue. This symmetry gives rise to two of the central tasks in communication: determining what parts of one's knowledge to share and what additional knowledge needs to be elicited from the other participant(s)~\citep{grice1975logic,lin2024decision}. Second, this style of evaluation requires a faithful simulator of human users. This is a challenging task in its own right because real human speakers may enter a conversation with only a vague sense of their own goals, preferences, and constraints~\citep{lintomlin2025usersim}.

\begin{figure}[t]
\centering
\input{tikz-figures/isotokens}
\vspace{-8pt}
\caption{\textbf{Isotoken evaluation}, showing task accuracy with a constant token budget sharded over varying numbers of turns. In most cases, performance is flat or decreasing as the turn budget increases, indicating that the models fail to exploit interactivity.}
\label{fig:isotokens}
\end{figure}

To measure the ability of language models to address the fundamental communicative challenges of multi-turn dialogue, we present a new form of multi-turn evaluation through collaborative \emph{private information games} (PINGs). The key feature is that each participant has access to private information, some of which is essential to the successful completion of the task. The participants must then collaboratively solve the task by building a shared understanding of their private information through linguistic communication.  To prevent participants from simply giving away their private information, we use information modalities that are not easy to convert into text (under reasonable length constraints): images and structured knowledge. This basic idea echoes much earlier work on ``information gap'' tasks in second-language instruction, in which language learners must share private information to jointly accomplish a goal~\citep{doughty1986information}.
We collect these private information games into a multi-turn benchmark called \benchmark. To enable comparative analysis at scale, we design the games to enable automatic verification and generation of new instances. This makes it possible to evaluate the multi-turn capabilities of language models from a new perspective that focuses on their proactive conversational capabilities. 

Success or failure in multi-turn private information games depends on two capabilities: (1) interacting effectively across multi-turn dialogues; and (2) reasoning about the game itself. Our primary interest is in the interactive capability; to isolate it, we propose a novel form of multi-turn scaling analysis, in which we fix the total number of tokens that can be exchanged in the interaction, and vary the number of turns (see \Cref{sec:isotoken-evals}). The performance with two turns is then a baseline, because if given a larger number of turns, the players can achieve the same communication by ignoring each other and sharding their message from the two-turn variant into however many turns are permitted. In well-designed tasks (see formal characterization in \Cref{sec:formal}), the players can do better by refining their communication based on the dialogue context. However, we observe that contemporary language models generally fail to achieve this. In most cases, there is no improvement as the turn budget grows, showing that the language model players are unable to use interaction to improve their task performance. In several cases, performance significantly \emph{decreases} with higher turn budgets, because the players prematurely terminate the dialogue rather than verifying their conclusions~\citep[][show a related finding in practical deployment scenarios]{cemri2025multi}. These negative results hold broadly across model scales and thinking budgets, suggesting that there is still significant room for improvement in multi-turn communication. 

\section{Levels of interactivity}
\label{sec:formal}
To formalize intuitions about the role of private information in multi-turn communication, we propose a framework for quantifying the \emph{level of interactivity} for a collaborative private information game. 
Consider a two-player game, where player~1 receives private information $X_1$ and player~2 receives private information $X_2$, which is at least partially disjoint. 
The shared goal is for player~1 to give an answer that gets a high score. (For this theoretical formulation it is convenient to allow only player~1 to give an answer.) Assume each player has access to an encoding function $f_i \in \mathcal{F}_i^n$, which maps the dialogue history and their private information $X_i$ to a string of length $\le n$. Then for any $(n, c)$, a game is level-$k$ interactive if there exist encoding functions $f_1 \in \mathcal{F}_1^n, f_2 \in \mathcal{F}_2^n$ such that by application of these encoding functions over $k$ turns of dialogue, player 1 can achieve an expected score $> c$. For example, a game is level-0 interactive if player 1 can answer without player 2's help; a game is level-1 interactive if there is an efficient encoding of $X_2$ such that the game is level-0 interactive; and so on. Clearly, games that are level-$k$ interactive are also level-$(k+1)$ interactive; \textbf{we seek games such that for some $k\ge 2$, the game is level-$k$ interactive but not level-$(k-1)$ interactive}, for reasonable values of $n$ and $c$. A related property is the game should be solvable by a single player with access to both $X_1$ and $X_2$. For a more formal discussion of levels of interactivity, with examples, see \Cref{sec:levels-of-interactivity}.

\section{Private information games}
\label{sec:tasks}
The \benchmark{} tasks aim for high levels of interactivity by giving each player private information that is difficult to transmit efficiently as text. 
The games are chosen to cover a few distinct forms of private data and modes of interaction, including images and structured data.
Some games are perfectly symmetric between the players (chess, COVR, name-game), while others assign distinct roles (MD3 and tangram). 
We now describe each game in detail. Full prompts are shown in \Cref{app:prompts}.

\begin{figure}[t]
    \centering
\input{examples/chess-pro}
\vspace{-8pt}
    \caption{Successful chess dialogue from Gemini 2.5 Pro with thinking. Chess is the sole task in which the models regularly exceed the token budget, as seen in turns 1, 5, 6, and 7.}
    \label{fig:chess-example-pro}
\end{figure}

\textbf{Chess.}
In the chess task, each player receives a board configuration in ASCII form. The boards are from the same game, and the players must determine which board is earlier in the game. The boards are chosen to be several moves apart, so that this should almost always be unambiguous. All moves are selected randomly from the uniform distribution over legal moves at each board position. Either player can end the game by saying \_YOURS\_ or \_MINE\_. An example is shown in \Cref{fig:chess-example-pro}.

The task has a baseline success rate of 50\%. An effective strategy would be to focus on two aggregate statistics: the number of pieces on the board and the rank of the most advanced pawns. Because these statistics are almost always monotonic over the course of a chess game, they could quickly identify the earlier board in most cases. The strategy of counting pieces is used by Gemini 2.5 Pro (see \Cref{fig:chess-example-pro}), but is not discovered by the other models that we tested (Gemini 2.5 Flash, GPT-4o, Qwen-VL8B, and Gemma3-12B). 

\begin{figure}[t]
    \centering
    \begin{examplebox}[top=1pt, bottom=-3pt]
            \scriptsize\sffamily
        \begin{tabularx}{\linewidth}{@{} p{0.15\linewidth}p{0.15\linewidth} X @{}}
        \vspace{0pt} %
        
            \begin{minipage}{\linewidth}
            \textbf{Alice}\\
            \includegraphics[width=.9\linewidth]{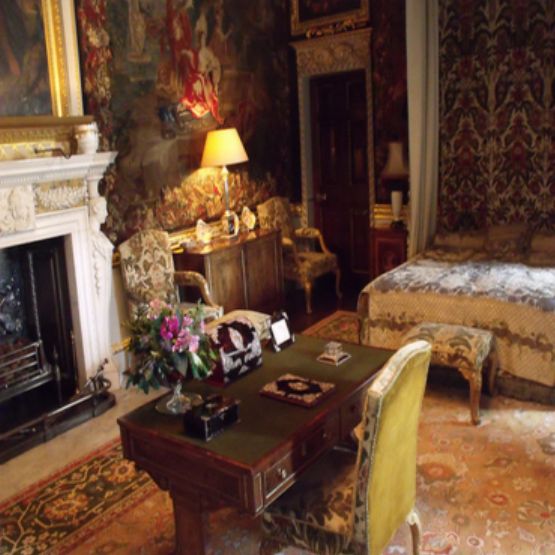}
            \vspace{1em}
            \end{minipage}
&
        \vspace{0pt} %

            \begin{minipage}{\linewidth}
            \textbf{Bob}\\
            \includegraphics[width=.9\linewidth]{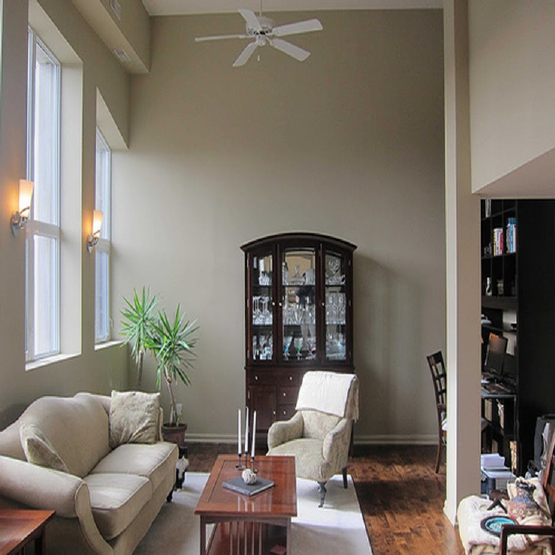}
            \end{minipage}
        &
        \vspace{0pt} %

            \textbf{QUESTION:} Is there both a bowl that is on a table that is in a bedroom and a paper towel that is on a table that is in a cafeteria?

            \begin{enumerate}[leftmargin=*,itemsep=0pt,parsep=0pt]
                \item \textbf{Alice:} Okay, let's figure this out! I see a bedroom with a desk and a fireplace. Do you see a cafeteria in your image?
                \item \textbf{Bob:} No, I don't see a cafeteria. I see a living room with a display cabinet.
                \item \textbf{Alice:} Okay. Does your living room have a table with a bowl on it?
                \item \textbf{Bob:} No, I don't see a bowl on a table. I do see a coffee table.
                \item \textbf{Alice:} Does your living room have a table with a paper towel on it?
                \item \textbf{Bob:} No, I don't see a paper towel on any table.
                \item \textbf{Alice:} Okay, so I see a bedroom, you see a living room. Neither of us has a bowl/paper towel on a table in the right places.
                \item \textbf{Bob:} ANSWER: False
            \end{enumerate}
        \end{tabularx}
\end{examplebox}

\vspace{-8pt}
    \caption{An example of a successful COVR dialogue from Gemma3-12B.}
    \label{fig:covr-example}
\end{figure}

\textbf{COVR.}
Multimodal reasoning benchmarks such as NLVR~\citep{suhr2017corpus} and COVR~\citep{bogin2021covr} require answering natural questions or evaluating sentences about sets of images. These benchmarks can easily be adapted to conversational evaluation by showing one image to each player and requiring them to communicate about their images so as to arrive at an answer. We use data from the COVR benchmark, restricting ourselves to the examples with exactly two images; the generalization to multi-party conversations about larger numbers of images is left for future work. An example is shown in \Cref{fig:covr-example}. As shown in the prompt (\Cref{tab:covr-prompt}), it was necessary to instruct the language models on the desired answer format to enable exact match evaluation.

\begin{figure}[t]
    \centering
\begin{examplebox}[top=-8pt, bottom=-10pt]
    \begin{tabularx}{\linewidth}{@{} p{0.45\linewidth} X @{}}
        
        \vspace{0pt} %
        \includegraphics[width=\linewidth]{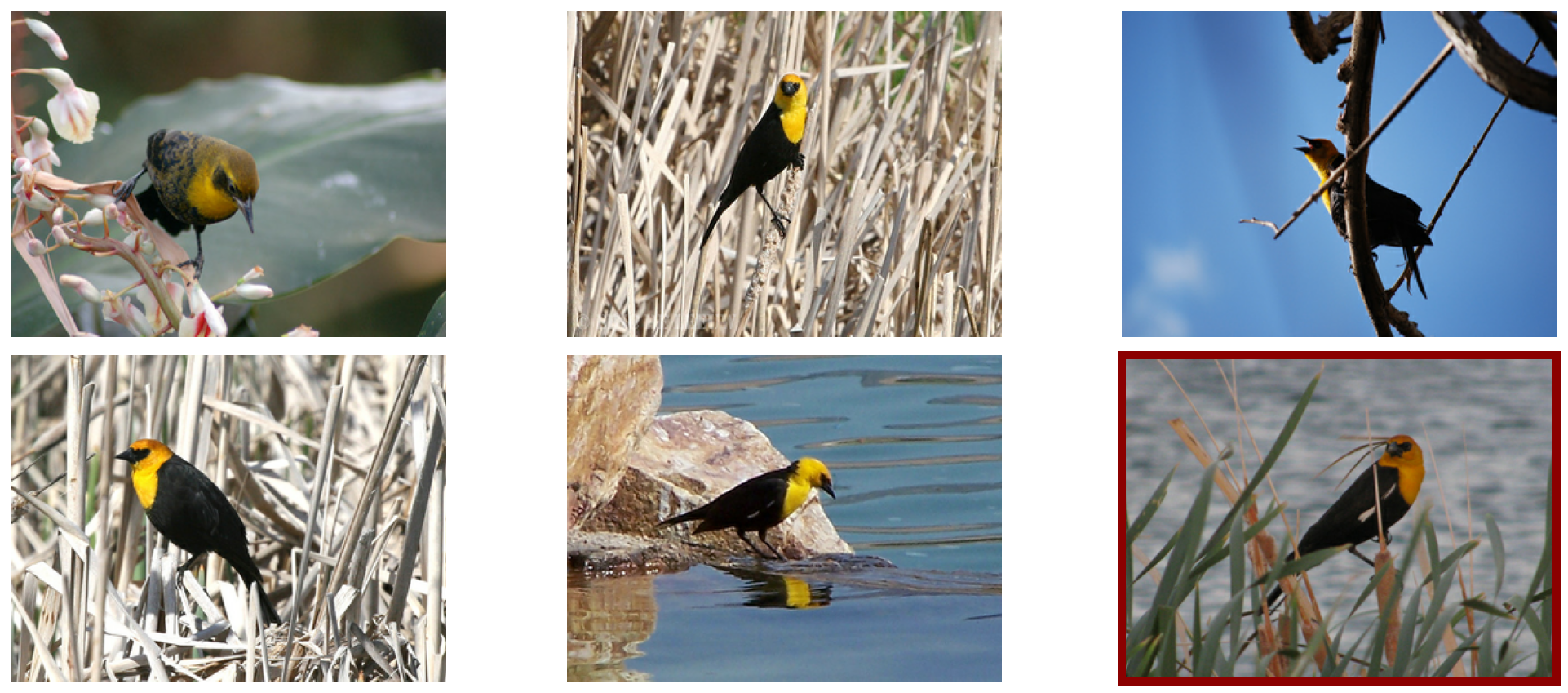} 
        
        & %
        
        \vspace{0pt} %
        \scriptsize\sffamily
        \begin{enumerate}[parsep=0em,itemsep=0pt, leftmargin=*]
            \item \textbf{Describer}: A bird with a bright yellow head and black body perched among tall reeds near water.
            \item \textbf{Guesser}: Can you describe the bird's pose or its surroundings further?
            \item \textbf{Describer}: The bird is facing left, gripping reeds with its beak, with a rippling water background.
            \item \textbf{Guesser}: ANSWER: Image 5
        \end{enumerate}
        
    \end{tabularx}
\end{examplebox}

    \begin{examplebox}[top=-8pt, bottom=-4pt]
        \begin{tabularx}{\linewidth}{@{} p{0.6\linewidth} X @{}}
        
        \vspace{0pt} %
        \includegraphics[width=\linewidth]{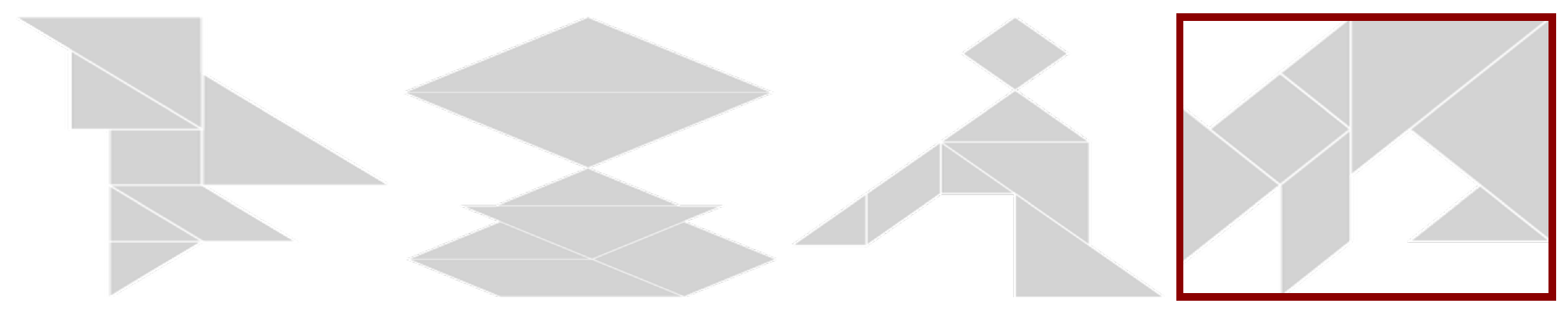}        
&
            \scriptsize\sffamily
            \begin{enumerate}[leftmargin=*]
                \item \textbf{Describer}: Describer: Black and white geometric pattern with diagonal stripes.
                \item \textbf{Guesser}: ANSWER: Image 0
            \end{enumerate}
 \end{tabularx}
           \end{examplebox}

\vspace{-8pt}
\caption{Top: example of successful dialogue for GPT-4o on the MD3 image selection task; the guesser sees all six images, while the describer sees only the highlighted image on the lower right. Bottom: unsuccessful dialogue for Qwen-VL8B on the Tangram task; the guesser sees all four images, while the describer sees  the highlighted image on the far right.}
    \label{fig:md3-example}
\end{figure}

\textbf{Image selection: MD3 and Tangram.}
\label{sec:tasks-image-selection}
The MD3 and tangram tasks are asymmetric image selection tasks. One player (the "describer") receives a single image and the other (the "guesser") receives a set of six photographs (MD3) or four geometrical shapes (tangram). The guesser must identify which of its images is the one seen by the describer, or to indicate if there is no match. This can be done by the describer enumerating features of their image, such as the focal object, the background, and visual features of the image's components. The guesser can ask questions that distinguish likely candidates. These tasks are derived from prior work on descriptions of geometric objects~\citep{ji2022abstract} and human-human dialogue~\citep{eisenstein2023md3}. Examples of MD3 and tangram are shown in \Cref{fig:md3-example}.

\textbf{Name-game.}
\label{sec:tasks-name-game}
In the name-game task, each player receives a structured database of people, and the task is to find the single record that is common to both databases. To modulate difficulty, the size of the database and number of features can be changed; we include examples with 9, 16, and 25 records and five features. This task is inspired by~\citet{he2017learning}, although we generated new data using Gemini 2.5 Pro. An example of this task is shown in \Cref{fig:name-game-example}. The task is quite challenging, and informally we have found that humans have a hard time identifying a good strategy.

\section{Isotoken multi-turn scaling evaluation}
\label{sec:isotoken-evals}

To evaluate the multi-turn scaling capabilities of language models, we propose a new form of evaluation, called \emph{isotoken evaluation}, which is motivated by the theory of levels of interactivity that was introduced in \Cref{sec:formal}. The core idea is that both players are given a fixed total token budget, and we vary the number of turns into which that budget must be partitioned. For example, a budget of 128 tokens per player (256 tokens in total) may be partitioned into 16 tokens per turn in a 16-turn setting, or 128 tokens per turn in a 2-turn setting. The intuition is that any dialogue that can be performed in $t$ turns can also be performed in $>t$ turns, because the players can ignore some of the turn boundaries. Thus, the task success should never decrease as the number of turns is increased, and for highly interactive tasks, we should expect task success to improve as the turn budget grows.\footnote{For comprehensive evaluations across all combinations of token and turn budgets, see \Cref{app:non-isotoken}.}

Both the token and turn budgets are upper limits: we do not force the models to use all available turns, as part of the challenge is to determine when the task has been successfully solved. For simplicity of implementation, we do not allow the models to partition the token budget arbitrarily, but rather allow exactly $\lfloor T/t \rfloor$ tokens per turn with a budget of $T$ tokens and $t$ turns. The token limit is given in the prompt, and all  models were able to consistently stay under budget, except in the chess task.\footnote{The token budget is implemented in subword tokens, but the language models appear to be capable of reasoning only about whitespace-delimited tokens: for example, when in thinking mode, Gemini 2.5 Pro and Flash explicitly count out the number of whitespace-delimited tokens in each proposed response before responding. As there are approximately 1.3 subword tokens per whitespace-delimited token, we have the prompt state a token budget that is $1.3 \times$ smaller than the true subword token budget. This may partly explain the difficulty in the chess task, where the number of subwords per whitespace token would be higher for chess-specific board position terminology like `A8'.} In the rare cases where the token budget was violated, we truncate the output. The models sometimes disregarded the turn budget, failing to complete the task on the final turn. This problem was cured by adding reminders in the prompts for the final and penultimate turns, as shown in the bottom of \Cref{tab:name-game-prompt}. 

We apply isotoken evaluation to the \benchmark{} tasks described in \Cref{sec:tasks}, with budgets of $\{2, 4, 8, 16\}$ turns and 128 tokens per player. We generate 1000 examples per task and evaluate five models: Gemini 2.5 Pro, Gemini 2.5 Flash, GPT-4o, Qwen-vl8B, and Gemma3-12B. The Gemini models are evaluated with and without thinking. (Thinking does not count against the token budget because the goal is a comparison that is \emph{communication}-matched, not \emph{compute}-matched.) All models use the same prompts, shown in \Cref{app:prompts}. 

\begin{figure}
\centering
\includegraphics[width=\textwidth]{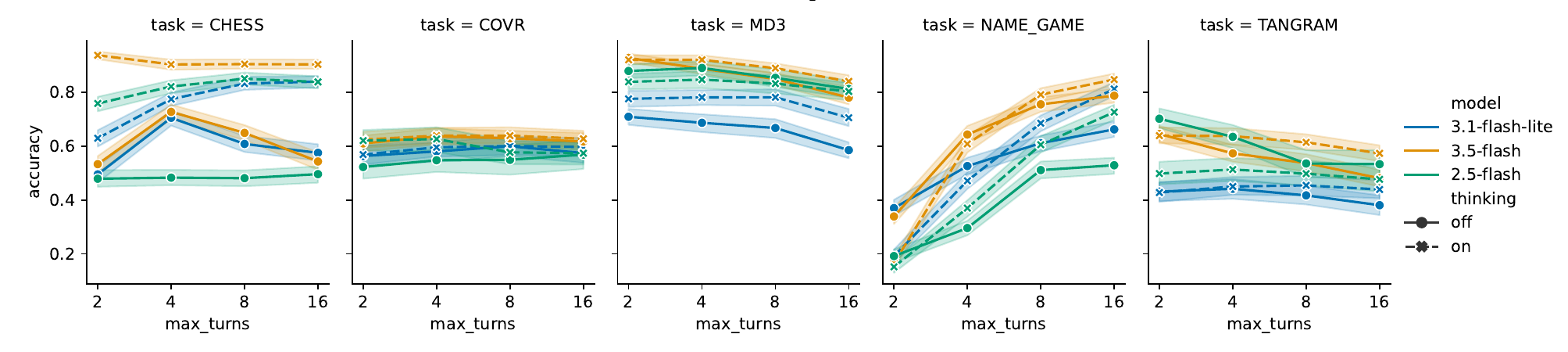}
\vspace{-24pt}
\caption{Isotoken evaluation of Gemini Flash models}
\label{fig:new-models}
\end{figure}

\textbf{Summary of results.} The main results are shown in \Cref{fig:isotokens}. With few exceptions, performance does not improve as the models are allowed more turns. In the image selection tasks, performance clearly gets worse as the turn budget increases. \Cref{fig:new-models} applies this evaluation to newer Gemini models, which show similar trends.
We now describe some qualitative differences in the scaling trends across tasks, along with discussion of how the models approach the tasks.
 
\textbf{Chess.}
In the chess task, the thinking models are clearly superior, and the non-thinking Gemini models improve somewhat with the number of turns. The remaining models perform at around the chance level of 50\%. As noted above, on this task it is possible to give a highly informative summary of a board position in a very small number of tokens, by listing the number of pieces per player and the rank of the most advanced pawns (with few exceptions, these indicators move monotonically over the course of a game). This is shown in \Cref{fig:chess-example-pro}, where both models report on the number of pieces on the board. Gemini 2.5 Pro is the only model to use this strategy, and it sometimes makes mistakes in counting. Other models tend to list apparently random subsets of their board positions before summarizing the situation and guessing at a conclusion (e.g., \Cref{fig:chess-sycophancy-example}).

\textbf{COVR.}
On the COVR task, which focuses on image analysis, the accuracies for all models are relatively flat as the number of turns increases. For the strongest model, Gemini~2.5 Pro (thinking), performance decreases monotonically from an overall best of 61.5\% accuracy at two turns, to 58.8\% at 16 turns. However, the non-thinking version of the same model improves its performance from 53.2\% to 59.3\% from two to 16 turns, suggesting that thinking and interactive dialogue may in some cases play similar roles. Most of the other differences in performance across turn budgets are small and statistically insignificant.

\textbf{Image selection: MD3 and Tangram.}
The image selection tasks show the strongest evidence of inverse scaling with the number of turns: all models suffer from reduced performance as the number of turns increases and the number of tokens per turn decreases. Any 2-turn dialogue could in principle be played over a larger number of turns, so there is nothing that makes the many-turn / few-token setting inherently more challenging. Some models waste tokens in each turn with uninformative introductions, but the main issue is that they do not use their full turn budgets, to a far greater extent than in the other tasks (\Cref{fig:isotoken-token-and-turns}). For example, on MD3, even Gemini 2.5 Pro uses fewer than four turns on average in the setting where it is permitted 16 turns and 16 tokens per turn. As a result, the total number of tokens used by each player  decreases as the number of turns grows (\Cref{fig:isotoken-token-and-turns}). This is despite our efforts to tune the prompts to encourage the models not to terminate the dialogue until completely sure of their answer (\Cref{tab:image-selection-prompt}). Premature  termination has been identified as a  cause of failures in practical multi-LLM applications~\citep{cemri2025multi}.

\begin{figure}[t]
\centering
\begin{examplebox}%
\scriptsize\sffamily
\begin{tabularx}{\linewidth}{@{} p{0.48\linewidth} X @{}}
    
    \begin{minipage}[t]{\linewidth}
        \vspace{0pt} %
        \textbf{Alice} \\
        \resizebox{\linewidth}{!}{%
            \begin{tabular}{|r|l|l|l|l|l|}
            \hline
            row & name & astrological sign & company & favorite musician & allergies \\
            \hline
            1 & Michael & Scorpio & Cisco & John Coltrane & Corn \\
            2 & Andrew & Aries & Amazon & Count Basie & Shellfish \\
            3 & Ethan & Cancer & Tesla & Miles Davis & Eggs \\
            4 & Asher & Aries & Sony & Ella Fitzgerald & Corn \\
            5 & Lucas & Capricorn & Amazon & Art Blakey & Fish \\
            6 & Charlotte & Libra & Cisco & Nina Simone & Mustard \\
            7 & Isabella & Leo & Tesla & Miles Davis & Eggs \\
            8 & Asher & Cancer & Google & Ella Fitzgerald & Corn \\
            9 & Isabella & Leo & HP & Miles Davis & Eggs \\
            \hline
            \end{tabular}%
        }
    \end{minipage}
    &

    \begin{minipage}[t]{\linewidth}
        \vspace{0pt} %
        \textbf{Bob} \\
        \resizebox{\linewidth}{!}{%
            \begin{tabular}{|r|l|l|l|l|l|}
            \hline
            row & name & astrological sign & company & favorite musician & allergies \\
            \hline
            1 & Isabella & Leo & Tesla & Miles Davis & Eggs \\
            2 & Lily & Sagittarius & HP & Herbie Hancock & Soy \\
            3 & Chloe & Pisces & Oracle & Herbie Hancock & Mustard \\
            4 & Chloe & Gemini & Qualcomm & Louis Armstrong & Gelatin \\
            5 & Sophia & Libra & Samsung & Wes Montgomery & Peanuts \\
            6 & Aurora & Capricorn & Apple & Billie Holiday & Sesame \\
            7 & Sofia & Leo & Amazon & Art Blakey & Tree Nuts \\
            8 & Ella & Pisces & Apple & Thelonious Monk & Tree Nuts \\
            9 & Samuel & Leo & Tesla & Miles Davis & Corn \\
            \hline
            \end{tabular}%
        }
    \end{minipage}
    \\
    \addlinespace %
    
    \begin{minipage}[t]{\linewidth}
        \vspace{0pt}
        \begin{enumerate}[parsep=0em, itemsep=0pt, topsep=0pt, leftmargin=*]
            \item \textbf{Alice:} Hello! I know a Michael who works at Cisco. Do you know him?
            \item \textbf{Bob:} I don't know a Michael at Cisco. Do you know anyone who works at Apple?
            \item \textbf{Alice:} I don't know anyone at Apple. I know an Andrew who works at Amazon. Do you know him?
            \item \textbf{Bob:} I don't know Andrew. Do you know anyone who is a Leo?
        \end{enumerate}
    \end{minipage}
    &
    
    \begin{minipage}[t]{\linewidth}
        \vspace{0pt}
        \begin{enumerate}[leftmargin=*, start=5]
            \item \textbf{Alice:} Yes, I know two Leos. One works at Tesla. Do you know an Ethan who works at Tesla and likes Miles Davis?
            \item \textbf{Bob:} I don't know Ethan. I know two Leos at Tesla liking Miles Davis, but different names. Do you know a Pisces?
            \item \textbf{Alice:} I don't know a Pisces. I know one Leo at Tesla liking Miles Davis. Her name is Isabella. Do you know her?
            \item \textbf{Bob:} SELECT ROW 1
        \end{enumerate}
    \end{minipage}
    
\end{tabularx}
\end{examplebox}

    \vspace{-8pt}
    \caption{A successful but lucky dialogue for Gemini 2.5 Flash (thinking) on the name-game.}
    \label{fig:name-game-example}
\end{figure}

\textbf{Name-game.}
\label{sec:results-name-game}
Name-game stands out as the task in which performance improves substantially with the number of turns. However, inspection of the transcripts shows that in many cases, the players simply guess randomly until they find a lucky match or run out of turns. This can be seen in \Cref{fig:name-game-example}, where Alice happens to mention Isabella in the final turn, and Bob marks this as a match without confirming all of the attributes. The example also highlights the questionable pragmatics of even relatively strong models: Alice's response in turn 5 strongly implies that Ethan is the astrological Leo who works at Tesla and likes Miles Davis, but in fact she has changed the topic and is no longer talking about Leos.

\begin{wrapfigure}{r}{0.4\textwidth}
\centering
\vspace{-12pt}
\definecolor{accessiblered}{HTML}{D55E00}
\definecolor{accessibleorange}{HTML}{E69F00}
\definecolor{accessiblesky}{HTML}{000000}
\definecolor{accessiblegreen}{HTML}{009E73}
\definecolor{accessibleblue}{HTML}{0072B2}
\definecolor{accessiblepurple}{HTML}{CC79A7}
\definecolor{accessibleblack}{HTML}{000000}

\begin{tikzpicture}
\begin{groupplot}[group style={group name=myplots, group size=2 by 1, horizontal sep=1.5cm, vertical sep=2cm, xlabels at=edge bottom, ylabels at=edge left},width=3.5cm,height=3cm,tick label style={font=\scriptsize},label style={font=\scriptsize},title style={font=\scriptsize},xmin=2,xmax=16,xmode=log,log basis x=2,xtick={2,4,8,16},xticklabels={2,4,8,16},ymin=0,ymax=1,grid=major,xlabel={max\_turns},ylabel={accuracy}]
    \nextgroupplot[
    title={
    }]
        \addplot [fill=accessiblered, opacity=0.15, draw=none, forget plot] coordinates { (2,0.228) (4,0.389) (8,0.541) (16,0.630) (16,0.688) (8,0.603) (4,0.451) (2,0.282) } -- cycle;
        \addplot [color=accessiblered, solid, mark=*, mark options={scale=0.8, solid}] coordinates { (2,0.255) (4,0.420) (8,0.572) (16,0.659) };
        \addplot [fill=accessiblered, opacity=0.15, draw=none, forget plot] coordinates { (2,0.210) (4,0.491) (8,0.691) (16,0.820) (16,0.866) (8,0.747) (4,0.553) (2,0.262) } -- cycle;
        \addplot [color=accessiblered, dashed, mark=*, mark options={scale=0.8, solid}] coordinates { (2,0.236) (4,0.522) (8,0.719) (16,0.843) };
        \addplot [fill=accessibleorange, opacity=0.15, draw=none, forget plot] coordinates { (2,0.169) (4,0.275) (8,0.469) (16,0.518) (16,0.580) (8,0.531) (4,0.331) (2,0.219) } -- cycle;
        \addplot [color=accessibleorange, solid, mark=square*, mark options={scale=0.8, solid}] coordinates { (2,0.194) (4,0.303) (8,0.500) (16,0.549) };
        \addplot [fill=accessibleorange, opacity=0.15, draw=none, forget plot] coordinates { (2,0.210) (4,0.362) (8,0.531) (16,0.694) (16,0.750) (8,0.593) (4,0.422) (2,0.262) } -- cycle;
        \addplot [color=accessibleorange, dashed, mark=square*, mark options={scale=0.8, solid}] coordinates { (2,0.236) (4,0.392) (8,0.562) (16,0.722) };
        \addplot [fill=accessiblesky, opacity=0.15, draw=none, forget plot] coordinates { (2,0.046) (4,0.186) (8,0.393) (16,0.722) (16,0.776) (8,0.455) (4,0.236) (2,0.076) } -- cycle;
        \addplot [color=accessiblesky, solid, mark=triangle*, mark options={scale=0.8, solid}] coordinates { (2,0.061) (4,0.211) (8,0.424) (16,0.749) };
    \nextgroupplot[hide axis, title={}, clip=false]
    \node[anchor=center, inner sep=5pt] at (axis description cs:0.5,0.5) {
      \scriptsize
      \renewcommand{\arraystretch}{0.9}
      \setlength{\tabcolsep}{3pt}
      \begin{tabular}{@{}l@{}}
        \textbf{model} \\
        \tikz[baseline=-0.5ex, x=1cm, y=1cm]\draw[accessiblered, thick] (0,0) -- (0.5,0) plot[mark=*, mark options={scale=0.8, solid}] coordinates {(0.25,0)};~Gemini 2.5 Pro \\
        \tikz[baseline=-0.5ex, x=1cm, y=1cm]\draw[accessibleorange, thick] (0,0) -- (0.5,0) plot[mark=square*, mark options={scale=0.8, solid}] coordinates {(0.25,0)};~Gemini 2.5 Flash \\
        \tikz[baseline=-0.5ex, x=1cm, y=1cm]\draw[accessiblesky, thick] (0,0) -- (0.5,0) plot[mark=triangle*, mark options={scale=0.8, solid}] coordinates {(0.25,0)};~Guess one baseline \\
        \\
        \textbf{thinking} \\
        \tikz[baseline=-0.5ex, x=1cm, y=1cm]\draw[black, solid, thick] (0,0) -- (0.5,0);~off \\
        \tikz[baseline=-0.5ex, x=1cm, y=1cm]\draw[black, dashed, thick] (0,0) -- (0.5,0);~on \\
      \end{tabular}
    };
\end{groupplot}
\end{tikzpicture}
\vspace{-18pt}
\caption{Comparison of the Gemini models against a simple guess-and-check heuristic on name-game.}
\label{fig:name-game-baseline}
\end{wrapfigure}

While the language models do improve their performance with the turn budget, this is driven primarily by poor performance at low turn budgets rather than effective use of interaction at high turn budgets. The models usually guess only a single row per turn, and therefore do not use all allowed tokens in the 2-turn and 4-turn cases. Indeed, a baseline that simply proposes a random record in each turn (sampled without replacement) displays a similar scaling trend, achieving a higher success rate than all models except Gemini 2.5 Pro (thinking) in the 16-turn setting (\Cref{fig:name-game-baseline}). By the nature of the task, this baseline's performance will improve rapidly with the turn limit. This explains much of the improvement in performance of the language models, which follow a similar strategy. 

\section{Linguistic analysis}
\label{sec:analysis}
To better understand the  performance of language models on \benchmark, we explore three possible explanations in the linguistic features of model output:
\begin{itemize}[itemsep=0pt]
\item \textbf{sycophancy} could cause LLMs to disregard their own private information, leading to collaborative failure when their private information is in fact crucial;
\item low \textbf{information density} might cause LLMs to struggle under token constraints;
\item lack of \textbf{goal-directedness} might cause LLMs to fail to elicit necessary information from the other player.
\end{itemize}
We first present linguistic analyses of the model outputs (\S\ref{sec:ling-features}), and then offer a rough comparison with similar human data from the MD3 task (\S\ref{sec:ling-human}). We find strong evidence against the first two hypotheses: LLMs are in fact quite capable of rejecting proposals that are incompatible with their private information (\cref{fig:early-proposals}), and LLM dialogues are significantly more information-dense than comparable human dialogues (\cref{fig:coherence-density-scatter}). Goal directedness is not definitively excluded as an explanation, but it is suggestive that the most successful models (Gemini thinking models) are not the ones with the highest coherence.

\subsection{Features}
\label{sec:ling-features}
\textbf{Sycophancy.}
Sycophancy is a major concern about today's large language models~\citep{perez-etal-2023-discovering}, due in part to its potential effects on human-human interaction patterns~\citep{cheng2025sycophantic}. In the context of \benchmark, sycophantic language model players might be too willing to accept claims or ideas by the other player, potentially overriding private information that could help them achieve a more accurate answer~\citep{stengel2025teaching}. If so, this would suggest that language models are likely to do the same in dialogues with human interlocutors, which would significantly limit their utility. We tested for sycophancy in two ways: (1) focused analysis of how models handle proposed answer hypotheses in the chess task (where answers are binary and the player roles are symmetric), and (2) use of LLM-as-a-judge to detect uncritical agreement and validation of false premises. We found that while there is considerable evidence for ``stylistic'' sycophancy according to the LLM judge, the language models evaluate each other's answer hypotheses critically, and are far more likely to accept correct hypotheses. For more details, please see \Cref{sec:sycophancy-task-performance} and \ref{sec:stylistic-sycophancy}.

\textbf{Information density.}
The isotoken evaluation paradigm leans heavily on the capability of models to communicate \emph{efficiently} --- with high information density --- due to the evaluation's use of tight token constraints. On the other hand, more interactive dialogue naturally requires conversational scaffolding (e.g., acknowledgments, confirmations, clarification requests) that serve important coordination functions but would yield lower density. A natural question, then, is whether the models that achieve the greatest task success are those that maintain high levels of information density even in settings with many turns and few tokens per turn. To test this, we focus on \emph{lexical density} \citep{Ure1971,Halliday1985}, a widely used measure in linguistics and discourse analysis that captures how much informational content a text conveys by examining the proportion of lexical (content) words relative to the total number of words. We implement a lexical density metric that combines two complementary signals: (1) the \emph{content-word ratio}, which measures the proportion of content that excludes fillers, stopwords, and disfluencies; (2) the \emph{novelty} of each turn, as measured by TF-IDF~\citep{sparckjones1972idf}, which is widely used to measure informativeness of words in a corpus. We define lexical density as the product of these two quantities. However, we find limited support for the hypothesis that stronger models would maintain higher density. Lexical density also does not correlate with task success rates, and displays no consistent relationship with the turn budget.
\textbf{This discrepancy between communicative effort and task success suggests that current LLMs can generate content-rich utterances but struggle to deploy that content strategically in service of collaborative goals}. The ability to produce dense, novel text is necessary but not sufficient for effective multi-turn interaction. For more details, please see \Cref{sec:lexical-density}

\begin{figure}
\centering
\input{tikz-figures/isotoken_turns_and_tokens}
\vspace{-12pt}
\caption{Tokens and turns per dialogue.}
\label{fig:isotoken-token-and-turns}
\end{figure}

\textbf{Goal directedness and coherence.}
Goal-directed conversations are coherent in the sense of maintaining focus on the overarching goal of interaction. 
Are language models able to maintain coherence on the \benchmark{} tasks? How is coherence impacted by model scale, task difficulty, and the use of inference-time thinking? To answer these questions, we apply \emph{centering theory}, a discourse-level framework for modeling local coherence by tracking how entities are introduced, maintained, and shift across adjacent utterances \citep{grosz_joshi_weinstein_1995,walker_iida_cote_1994}. Historically, centering theory has played a number of roles; for our purpose, most critical is its role as a foundation for modeling coherence in text generation and evaluation, providing interpretable signals that correlate with human judgments of discourse quality~\citep{barzilay-lapata-2008-modeling,miltsakaki-kukich-2000-role}. The key property of centering theory is that it focuses on \emph{local transitions}, capturing how speakers manage continuity and change in what a discourse is ``about'' from one utterance to the next. This is implemented through the notion of \emph{centers}, discourse entities that are salient at a given point; more coherent discourses are more likely to feature utterance-to-utterance transitions that retain the same discourse elements in the central positions. We operationalize these principles into a heuristic measure of centering-based coherence, by tracking the centering transitions between adjacent utterances; for details on the method, please see \Cref{sec:coherence}. We find that in general among the non-thinking models, \textbf{the most powerful and effective models also produce the most coherent dialogues; however, surprisingly, thinking significantly reduces coherence} even as it improves performance on some of the \benchmark{} tasks. This may be because our lexico-syntactic measure of coherence combines multiple distinct phenomena. In some cases, the coherence score reflects the desired goal-directed focus of the participants. In other cases, it reflects relatively superficial features, such as when a turn ends with a question and the next turn begins with an explicit answer to that question, e.g. ``Do you know a Pisces? / I don't know a Pisces.'' 

\subsection{Comparison with human dialogues}
\label{sec:ling-human}
A closely-related image selection task is performed by humans in prior work~\citep{eisenstein2023md3}.\footnote{These transcripts are available at \url{https://www.kaggle.com/datasets/jacobeis99/md3en}} The image sets are the same, but there are a few important differences. The humans had no limit on the turns or tokens that they could use, but were asked to solve as many task instances as they could in five minutes. ``No match'' was not an option, but the number of distractor images was 12 rather than six. With those caveats in mind, it is possible to make a rough comparison between the properties of human-human dialogues with those of dialogues between language models~\citep{shaikh2024grounding}. 
While this comparison is imperfect, it represents a first step beyond prior work on evaluation of LLM multi-turn capabilities, which makes no comparison with human language at all~\citep[e.g.,][]{bai2024mt,kwan2024mt,laban2025llms}.

    \textbf{Task performance.}     
    Humans achieve task success rates between 86-91\%. This exceeds the task success rates for language models, although the success rates are not directly comparable for the reasons noted above.
    
    \textbf{Efficiency.} Humans averaged 10.6 whitespace-delimited tokens per turn (median $= 5$), and 5.5 turns per prompt (median $=4$). This means that on average humans used only 60 tokens per dialogue in total (across both players). The short turn lengths and relatively interactive dialogues may be linked to the speech modality~\citep{biber1991variation}. While the language models were \emph{allowed} 256 tokens per dialogue, they \emph{used} significantly fewer: as shown in \Cref{fig:isotoken-token-and-turns}, each player used fewer than 50 tokens in all settings, except GPT-4o at 2 turns. Thus, the language models were less token-efficient than the human players, while also failing to fully utilize their token budget to maximize task performance.  
    
    \textbf{Linguistic features.} Human dialogues had an average lexical density of $24.95$; this was lower than all of the language models on all versions of MD3, which ranged from 40--70. Again, a likely explanation is the speech modality: 89\% of the human dialogues contained a backchannel or filler token ("ok", "okay", "done", "um", "uh") and 6.8\% contain an apology.  
    The human dialogues have an average coherence score of~$0.86$, which exceeded most of the models in this task, except for Gemini~2.5 Pro (no thinking). 
    The relationship between lexical density and coherence on this task is shown in \Cref{fig:coherence-density-scatter}.

\begin{figure}
\centering
\begin{tikzpicture}
    \begin{groupplot}[
        group style={
            group size=2 by 1,
            horizontal sep=2cm,
            vertical sep=1cm,
        },
        width=6cm, height=3cm,
        ybar,
        /pgf/bar width=15pt,
        enlarge x limits=0.5,
        symbolic x coords={incorrect, correct},
        xtick=data,
        ymin=0,
            tick label style={font=\scriptsize},
            label style={font=\scriptsize},
            title style={font=\scriptsize},
        ylabel={count},
        xlabel={},
        ymax={350},           %
        enlarge y limits={upper, value=0.15}, %
        axis lines*=left,
        ymajorgrids=false,
        legend entries={rejected, accepted},
        legend to name=GlobalLegend, 
        legend style={draw=none, /tikz/every even column/.append style={column sep=0.2cm}},
        nodes near coords,  %
        every node near coord/.append style={font=\footnotesize}, %
        /pgf/number format/precision=0, %
    ]
    \nextgroupplot[title={Gemini 2.5 Flash}, ]
        \addplot[fill=white, draw=black] coordinates {(incorrect, 20.0) (correct, 18.0) };
        \addplot[fill=seabornGreen, draw=black] coordinates {(incorrect, 39.0) (correct, 353.0) };
    \nextgroupplot[title={Gemini 2.5 Pro}, ylabel={}, yticklabels={,},]
        \addplot[fill=white, draw=black] coordinates {(incorrect, 10.0) (correct, 9.0) };
        \addplot[fill=seabornGreen, draw=black] coordinates {(incorrect, 10.0) (correct, 187.0) };
    \end{groupplot}

    \node[anchor=north west] at (group c2r1.north east) [xshift=0.5cm] {
        \begin{tabular}{l} 
            \ref{GlobalLegend}
        \end{tabular}
    };
\end{tikzpicture}
\vspace{-18pt}
\caption{Counts of early answer proposals in the chess task for the Gemini~2.5 thinking models. Some incorrect proposals are accepted, but at a far lower rate than correct proposals.}
\label{fig:early-proposals}
\end{figure}

In summary, humans are able to perform the MD3 image selection task with high accuracy, despite low lexical density. This may be partly explained by the relatively high coherence of human dialogues, which exceeds all but one of the language models. Beyond coherence, the outstanding token efficiency of human dialogues suggests that they approach these tasks in a more goal-directed way than even the strongest language models.

\section{Related work}
Early multi-turn evaluations of large language models focused on manually-designed follow-up questions to elicit multi-turn conversations~\citep{zheng2023judging,kwan2024mt,bai2024mt}. \citet{laban2025llms} describe these as "episodic" evaluations, because each turn introduces a subtask that can in principle be evaluated in isolation. They propose an alternative evaluation in which single-turn questions are sharded into multiple turns, which they find to significantly diminish the performance of even the strong language models. However, the sharding is an arbitrary source of noise rather than the natural result of attempts to communicate  private information under information bandwidth constraints.

Self-play has been widely used to generate training and evaluation data for task-oriented  dialogue, typically involving a user simulator and an assistant~\citep{shah-etal-2018-bootstrapping,wang2023mint,li2024mediq,shao2024collaborative,dou2025simulatorarena}. Although this setup mirrors deployment, it is challenging for evaluation of the multi-turn collaboratively capability because user simulation requires faithfully reproducing the many foibles of human speakers, such as laziness, imprecision, and reliance on untested assumptions~\citep{lintomlin2025usersim}. 
Progress on this orthogonal capability of user simulation may alter multi-turn performance in unpredictable ways. As simulators improve, performance could become worse, because the simulators more faithfully imitate grounding failures and comprehension gaps. Alternatively, performance may appear to improve, because stronger user simulators may structure their communication more coherently. 
Private information games sidestep these issues: the language models do not need to simulate human users and are free to make their best efforts to achieve task success under the given constraints. 

\begin{figure}
\centering
\begin{tikzpicture}
\definecolor{colorFalse}{HTML}{006685} %
\definecolor{colorTrue}{HTML}{A10077}  %

\pgfplotstableread{
x y class_id model
51.28666666666667 0.7266666666666667 0 {GPT-4o}
39.14666666666667 0.71 1 {Gemini 2.5 Flash}
54.07333333333333 0.66 2 {Gemini 2.5 Flash, thinking}
45.45333333333334 1.0133333333333334 3 {Gemini 2.5 Pro}
52.75333333333333 0.61 4 {Gemini 2.5 Pro, thinking}
58.98666666666667 0.5 5 {Gemma3-12b}
69.16666666666667 0.5 6 {Qwen-VL8b}
}\datatable

\begin{axis}[
    width=12cm,
    height=3.7cm,
    xlabel={\footnotesize{lexical density}},
    ylabel={\footnotesize{coherence score}},
    label style={font=\small},
    xmin=20, xmax=75,
    ymin=0.43, ymax=1.05,
    grid=major,
    xtick={20,30,40,50,60,70},
    ytick={0.4,0.6,0.8,1.0},
    tick label style={font=\scriptsize},
]

\addplot[
    only marks,
    mark=*, mark size=3pt,
    color=colorFalse,
    forget plot,
    point meta=explicit symbolic, %
    nodes near coords,
    every node near coord/.style={
        anchor=west, 
        font=\sffamily\scriptsize, 
        text=black, 
        inner sep=3pt
    },
    restrict expr to domain={\thisrow{class_id}}{0:0}
] table [x=x, y=y, meta=model] {\datatable};
\addplot[
    only marks,
    mark=x, mark size=3.5pt, line width=1.5pt,
    color=colorFalse,
    forget plot,
    point meta=explicit symbolic, 
    nodes near coords,
    every node near coord/.style={
        anchor=east, 
        font=\sffamily\scriptsize, 
        text=black, 
        inner sep=3pt
    },
    restrict expr to domain={\thisrow{class_id}}{1:1}
] table [x=x, y=y, meta=model] {\datatable};
\addplot[
    only marks,
    mark=x, mark size=3.5pt, line width=1.5pt,
    color=colorTrue,
    forget plot,
    point meta=explicit symbolic, 
    nodes near coords,
    every node near coord/.style={
        anchor=west, 
        font=\sffamily\scriptsize, 
        text=black, 
        inner sep=3pt
    },
    restrict expr to domain={\thisrow{class_id}}{2:2}
] table [x=x, y=y, meta=model] {\datatable};
\addplot[
    only marks,
    mark=square*, mark size=2.5pt,
    color=colorFalse,
    forget plot,
    point meta=explicit symbolic, 
    nodes near coords,
    every node near coord/.style={
        anchor=west, 
        font=\sffamily\scriptsize, 
        text=black, 
        inner sep=3pt
    },
    restrict expr to domain={\thisrow{class_id}}{3:3}
] table [x=x, y=y, meta=model] {\datatable};
\addplot[
    only marks,
    mark=square*, mark size=2.5pt,
    color=colorTrue,
    forget plot,
    point meta=explicit symbolic, 
    nodes near coords,
    every node near coord/.style={
        anchor=east, 
        font=\sffamily\scriptsize, 
        text=black, 
        inner sep=3pt
    },
    restrict expr to domain={\thisrow{class_id}}{4:4}
] table [x=x, y=y, meta=model] {\datatable};
\addplot[
    only marks,
    mark=+, mark size=3.5pt, line width=1.5pt,
    color=colorFalse,
    forget plot,
    point meta=explicit symbolic, 
    nodes near coords,
    every node near coord/.style={
        anchor=east, 
        font=\sffamily\scriptsize, 
        text=black, 
        inner sep=3pt
    },
    restrict expr to domain={\thisrow{class_id}}{5:5}
] table [x=x, y=y, meta=model] {\datatable};
\addplot[
    only marks,
    mark=diamond*, mark size=3.5pt,
    color=colorFalse,
    forget plot,
    point meta=explicit symbolic, 
    nodes near coords,
    every node near coord/.style={
        anchor=south, 
        font=\sffamily\scriptsize, 
        text=black, 
        inner sep=3pt
    },
    restrict expr to domain={\thisrow{class_id}}{6:6}
] table [x=x, y=y, meta=model] {\datatable};

\addplot[only marks, mark=star, mark size=4.5pt, color=blue, forget plot] coordinates {(25.0, 0.86)};
\node[anchor=south west, align=left, font=\sffamily\scriptsize] at (axis cs:25.0,0.86) {human\\speech};
\end{axis}
\end{tikzpicture}
\vspace{-8pt}
\caption{Lexical density and coherence for AI models and humans on the MD3 image selection task. The language model scores are averaged over all turn budgets.}
\label{fig:coherence-density-scatter}
\end{figure}

Two papers take a similarly symmetrical approach to evaluation of collaboration between language model agents, finding significant challenges in multi-agent collaboration. \cite{davidson2025collaboration} design a navigation task in which each agent sees a partially-occluded version of the same map, and must communicate to identify a path. They find that this setting is significantly more challenging than one in which a single agent sees the whole map. In CooperBench, a set of coding tasks is divided between two agents~\citep{khatua2026cooperbench}, showing that agents fail to coordinate their work~\citep[see also][]{cemri2025multi}. Our approach focuses on simpler tasks that can be completed with a single, short output. This allows us to quantify the scaling of language model collaboration across varying turn budgets and to compare linguistic phenomena across diverse types of tasks.

Unconstrained self-play evaluation has also been applied in the chit-chat setting by \citet{ghandeharioun2019approximating}, who focus on measures of sentiment and semantic coherence across the dialogue~\citep[see also][]{gooding2025interaction}. We explore measures of coherence in \Cref{sec:coherence}, but rather than using word and sentence embeddings, we build on linguistic theories about the role of discourse entities through the conversation. Another distinction is that we consider dialogues in which the participants must share information to accomplish a task, rather than more open chitchat~\citep[see also][for a chitchat evaluation that focuses on whether LLM-generated turns can be detected in human-human dialogues]{duan2024botchat}.

\section{Conclusion}
We have presented \benchmark{}, a benchmark for evaluating multi-turn conversational capabilities through private information games. The isotoken evaluation methodology fixes the total communication budget while varying the number of turns, isolating multi-turn interaction capabilities from raw task-solving ability. This reveals a consistent pattern: language models generally fail to improve, and often get worse, as they are given more opportunities for interactive communication. 
Linguistic analysis provides some partial explanation: models exhibit stylistic sycophancy (e.g., spurious apologies and uncritical agreement), wasting tokens without advancing the task. And despite generating locally coherent dialogues, this coherence reflects formulaic patterns rather than strategic progress toward shared understanding. Comparison with human performance on the image guessing task highlights the gap: humans achieve higher accuracy with far fewer tokens, suggesting the challenge lies not in capability but in communication strategy --- knowing what to share, what to elicit, and when to commit.

Private information is a defining feature of real-world communication: we rarely know exactly what our interlocutors know, want, or need. \benchmark{} offers a rigorous framework for measuring progress on this fundamental challenge, and our results suggest that substantial advances in multi-turn interaction remain ahead.

\paragraph{Acknowledgments}
Thanks to Diego Antognini and Kevin Robinson for their assistance with engineering aspects of the project.
The paper benefited from feedback from Bernd Bohnet, Ryan Burnell, Elizabeth Clark, William Cohen, Chris Dyer, Asma Ghandeharioun, Anastasios Kement, Tal Linzen, Annie Louis, Claudiu Musat, and Hannah Rashkin.
Thanks also to the CoLM reviewers for several constructive suggestions that improved the paper.

\bibliography{main}
\bibliographystyle{Template-2026/colm2026_conference}

\clearpage
\appendix
\renewcommand\thesection{\Alph{section}}
\renewcommand{\thefigure}{\thesection.\arabic{figure}}
\renewcommand{\thetable}{\thesection.\arabic{table}}
\addcontentsline{toc}{chapter}{Appendices} %
\etocsetnexttocdepth{2}
\localtableofcontents
\clearpage
\section{Language model usage} 
We used language models in the following ways: (1) creating figures and tables by generating LaTeX code; (2) helping write data analysis code; (3) formatting bibtex entries; (4) automatically assessing sycophancy, as described in \Cref{sec:stylistic-sycophancy}; (5) generating the name-game examples, as described in \Cref{sec:tasks-name-game}; (6) reviewing the paper and suggesting areas in which the writing could be improved.
\section{Levels of interactivity}
\label{sec:levels-of-interactivity}
To formalize intuitions about the role of private information in multi-turn communication, we propose a framework for quantifying the \emph{level of interactivity} for a collaborative private information game. 
Consider a two-player game, where player~1 receives private information $X_1$ and player~2 receives private information $X_2$. 
The shared goal is for player~1 to give an answer that gets a high score. (For this theoretical formulation it is convenient to allow only player~1 to give an answer.)
The private information need not be completely disjoint: for example, both $X_1$ and $X_2$ may contain some description of the task, but they should also each contain at least some information that is unique. 

\newcommand{\eacc}[1]{\ensuremath \mathbb{E}[\textsc{score}(#1)]}
Assume each player~$i \in \{1, 2\}$ has an encoding function $f_i \in \mathcal{F}_i^n = \{ f: \mathcal{X}_i \times \mathcal{S} \to \mathcal{V}^n\}$, where $\mathcal{V}^n$ is the set of strings with $n$ tokens on vocabulary $\mathcal{V}$ and $\mathcal{S} = (\mathcal{V}^n)^*$ is the set of possible conversational histories. The history is then updated by concatenation, $S_t := S_{t+1} \circ f_{i(t)}(X_{i(t)}, S_{t+1})$, with $t$ indicating the number of remaining turns and $i(t)$ indicating the active player at turn $t$. We will use $i(t) = t \text{ mod } 2 + 1$, with $i(0)=1$ indicating that player 1 always provides the final answer.  At the initial turn $T$, we have $S_T = \varnothing$, and we will write $f(X)$ as shorthand for $f(X, S_T=\varnothing)$. Finally, let $\eacc{X_1, S} \ge 0$ represent the expected score from player $1$ answering based on private information $X_1$ and state $S$. We give concrete examples of such games and encoding functions immediately below.

For constants $n \in \mathbb{Z}_+, c \in \mathbb{R}_+$, we can characterize games by their level of interactivity:
\begin{itemize}[parsep=1ex] 
\item Level 0: The game is solvable without $X_2$, so no communication is needed: $\eacc{X_1, \varnothing} > c$.
\item Level 1: There is an efficient encoding function for $X_2$ that makes the game level-0 interactive: 
\begin{align*}
\exists f_2 \in \mathcal{F}_2^n : \, & \eacc{X_1, S_0} > c\\
\text{where } & S_0 = f_2(X_2).
\end{align*}
This means that the game is solvable in one round of conversation, with player~2 transmitting the length-restricted message $f_2(X_2)$ to player~1. For example, suppose that the game is image classification, so that $X_2$ is an image and $Y$ labels the class of the main object in the image. Player 2 does not need to know anything about $X_1$ (indeed, $X_1$ can even be empty) to transmit the message $Y=f_2(X_2)$ to player~1.
\item Level 2: There is an efficient encoding function for $X_1$ such that sharing this encoding makes the game level-1 interactive: 
\begin{align*}
\exists f_1 \in \mathcal{F}_1^n,\, \exists f_2 \in \mathcal{F}_2^n :\,& \eacc{X_1, S_0} > c\\
\text{where }& S_1 = f_1(X_1)\\
& S_0 = S_1 \circ f_2(X_2, S_1) = f_1(X_1) \circ f_2(X_2, f_1(X_1))
\end{align*}
For intuition, suppose that $X_2$ is an image and $X_1$ is another image and a text proposition (e.g., ``there are more beads in my image than your image'', $f_1(X_1)$ = ``how many beads'', $f_2(X_2, f_1(X_1)) = \text{"three"}$), and $Y$ labels whether the proposition is true. This example corresponds to our multiplayer implementation of COVR~\citep{bogin2021covr}, detailed in \Cref{sec:tasks}, although some instances from this task may not be resolvable in such a minimal dialogue.
\item Level 3: There is an efficient encoding of $X_2$ such that sharing this encoding makes the game level-2 interactive:
\begin{align*}
\exists f_1 \in \mathcal{F}_1^n, \, \exists f_2 \in \mathcal{F}_2^n : \, & \eacc{X_1, S_0} > c\\
\text{where }& S_2 = f_2(X_2)\\
& S_1 = S_2 \circ f_1(X_1, S_2) = f_2(X_2) \circ f_1(X_1, f_2(X_2))\\
& S_0 = S_1 \circ f_2(X_2, S_1) \\&\quad = f_2(X_2) \circ f_1(X_1, f_2(X_2)) \circ f_2(X_2, f_2(X_2) \circ f_1(X_1, f_2(X_2))
\end{align*}
It is harder to construct intuitive examples at this level and above. However, imagine the game of identifying elements that are shared between structured objects $X_1$ and $X_2$. The game can be solved in the following fashion: $f_2(X_2)$ lists at a high level the elements present in $X_2$; $f_1(X_1, S_2)$ identifies a subset of those elements that are also present in $X_1$; $f_2(X_2, S_1)$ describes the detailed \emph{features} of those elements in $X_2$; and given this information, player $1$ outputs a list of elements that have identical features in both $X_1$ and $X_2$.
\item Level $k>3$: There is an efficient encoding of $X_1$ or $X_2$ (depending on whether $k$ is even or odd) such that sharing this encoding makes the game level $k-1$ interactive.
\end{itemize}

Clearly games that are level-$k$ interactive are also level $(k+1)$-interactive. We seek games such that for some $k\ge 2$, the game is level-$k$ interactive but not level $(k-1)$-interactive, for reasonable values of $n$ and $c$. 
A related property is the game should be solvable by a single agent with access to both $X_1$ and $X_2$. 
Note that interactivity is a property of a game \emph{and model}, because a strong model might be able to solve a game in a smaller number of turns than a weaker model by producing better encodings or making better inferences. It is therefore not possible to analytically compute a game's level of interactivity. Rather, levels of interactivity are a conceptual tool to help formalize the design of our multi-turn evaluations.

\section{Additional statistics}
\label{app:non-isotoken}
\begin{figure}
\centering
\includegraphics[width=1.1\textwidth]{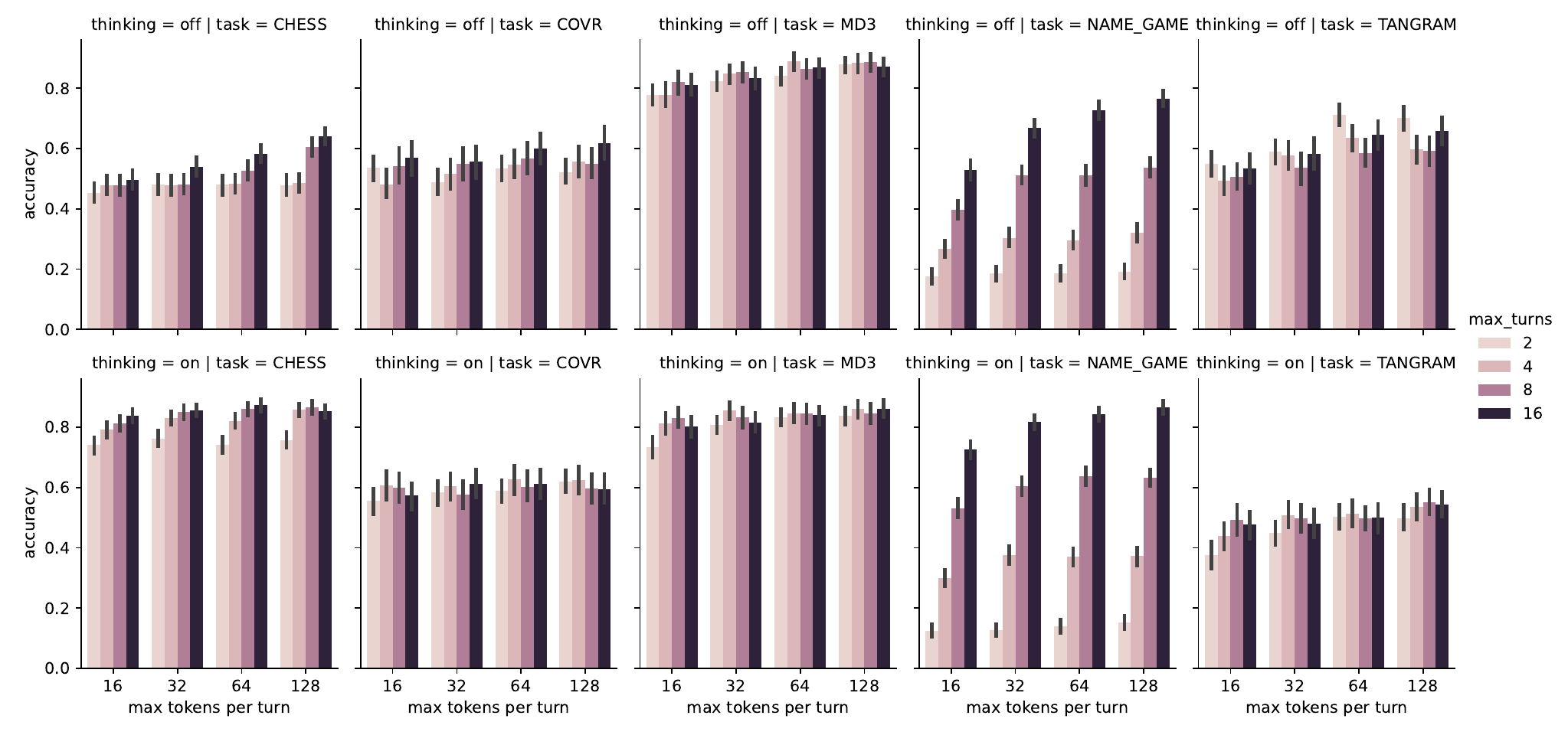}
\caption{Full turn/token budget sweep for Gemini 2.5 Flash}
\label{fig:full-sweep-2.5}
\end{figure}

\begin{figure}
\centering
\includegraphics[width=1.1\textwidth]{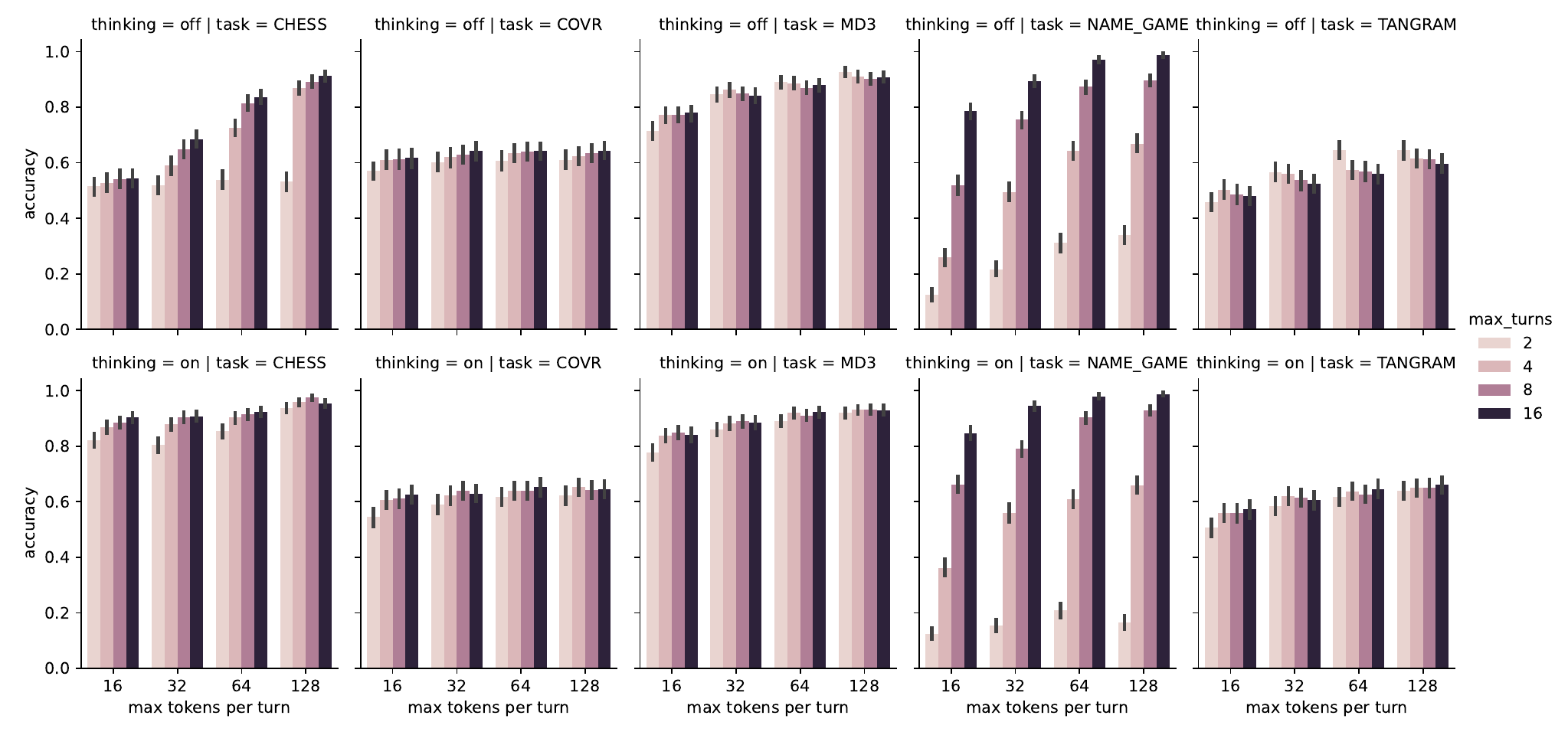}
\caption{Full turn/token budget sweep for Gemini 3.5 Flash}
\label{fig:full-sweep-3.5}
\end{figure}

In the isotoken evaluations, we hold the total token budget constant. We now present supplementary results for all combinations of turn and token budgets, for Gemini Flash versions 2.5 (\cref{fig:full-sweep-2.5}) and 3.5 (\cref{fig:full-sweep-3.5}) The results are grouped by tokens per turn.  At fixed token-per-turn budgets, increasing the turn budget increases the overall communication bandwidth. Even in this setting, increasing the turn budget does not yield consistent performance improvements in the MD3, TANGRAM, and COVR tasks. As before, performance increases with the number of turns in NAME-GAME, due to the "guess-one-row-per-turn" strategy described in \Cref{sec:isotoken-evals}. In CHESS, Gemini 3.5 Flash demonstrates improvements with the turn budget, but for Gemini 2.5 Flash this is contingent on the thinking strategy: without thinking, there are some performance improvements when the models are allowed a very large overall token budget ($\ge 1024$, which is 4x larger than the isotoken setting); with thinking, there are improvements over the 2-turn setting. We believe this is due to the increase in the total number of available tokens (constant tokens-per-turn, increasing number of turns), which allows the players to describe their boards more comprehensively.

\section{Private Information Game Prompts}
\label{app:prompts}
Prompt examples for all the \benchmark\ tasks are shown in Tables~\ref{tab:chess-prompt}, \ref{tab:covr-prompt}, \ref{tab:name-game-prompt}, and \ref{tab:image-selection-prompt}.

\newtcblisting{promptbox}{
    colback=white,
    colframe=black,
    arc=10pt,
    boxrule=1pt,
    listing only, %
    listing options={
    basicstyle=\ttfamily\tiny,
    breakindent=0pt,      %
        breaklines=true,      %
        columns=fullflexible, %
        gobble=0              %
    }
}

\begin{table}
    \begin{promptbox}
Your name is Bob and my name is Alice. You are a chess expert. We are going to talk about chess games using diagrams in which the white pieces are shown by upper-case letters like `K' and `Q', and the black pieces are shown by lower-case letters like `k' and `q'. Empty squares are shown as periods, `.'.

For example, in this board, White has moved its pawn to D4 and its knight to A3, while Black has moved its pawn to C5:

r n b q k b n r
p p . p p p p p
. . . . . . . .
. . p . . . . .
. . . P . . . .
N . . . . . . .
P P P . P P P P
R . B Q K B N R

Each of us will be shown a board that the other player cannot see. The two boards are from the same game of chess. Our task is to determine which board came earlier in the game. To accomplish this, we must describe our boards to each other.

You can use only 22 words per turn, so you cannot fully specify the locations of all pieces. Instead, describe the key features that will help us determine which board came first.

After discussing it, if you think your board came first, say `_MINE_'; if you think my board came first, say `_YOURS_'. Otherwise, you can continue to describe your board to me, or ask questions about my board.

It is important to get this right, so let's discuss until we are sure. However, if I say this is your final turn, then you **must** give a final answer that is either `_MINE_' or `_YOURS_'.

Here is your board, which I cannot see:

. n . . q . n r
r b . p k . b .
. . p . . . . .
p . . . P p p p
. . B P . . . .
. P . . . . P P
P . P N K P . .
R N . Q . R . .

# CONVERSATION

Here is our conversation history.

1. ME: Hello Bob. On my board, all black pieces are on their starting squares. White has moved the d-pawn to d4. No other white pieces

You have 7 turns left.

2. YOU: 
\end{promptbox}
\caption{Example chess prompt, after turn 1 of the dialogue.}
\label{tab:chess-prompt}
\end{table}

\begin{table}
    \begin{promptbox}
# INSTRUCTIONS

You are Alice, and you are playing a collaborative game that involves having a conversation with another player about two images. You can see the `left' image and the other player can see the `right' image.

Here is the `left' image: <start_of_image>

With the help of the other player, you have to answer the question:
"How many images contain at least 1 women that are watching child that is wearing helmet?"

To do this, you will have a conversation with the other player, for example by sharing information about your image or asking about the other player's image. In each turn you can use at most 22 words, so be concise!

When you are ready, output `ANSWER:' followed by your answer to the question. Your answer should be a *single* word, a number, or (rarely) a short phrase, e.g.:

`ANSWER: 4'
`ANSWER: True'
`ANSWER: next door'

If it is a true/false question, make sure to output `True' or `False'.
\end{promptbox}
\caption{Example COVR prompt, on the first turn of the dialogue.}
\label{tab:covr-prompt}
\end{table}

\begin{table}
\begin{promptbox}
# INSTRUCTIONS

Your name is Bob, and we are going to play a conversational game. You have a database of people who you know: 

```
row,name,company,favorite color,favorite musician,city
1,Noah,IBM,Mustard Yellow,Wes Montgomery,Chicago
2,Joshua,Samsung,Charcoal,Miles Davis,Sydney
3,James,HP,Charcoal,Charlie Parker,Rome
4,Madison,Google,Mustard Yellow,Thelonious Monk,Berlin
5,Elizabeth,Foxconn,Coral,John Coltrane,Rio de Janeiro
6,Asher,Amazon,Teal,John Coltrane,Moscow
7,Asher,Foxconn,Teal,John Coltrane,Moscow
8,Owen,Sony,Peach,Wes Montgomery,London
9,Madison,Foxconn,Coral,John Coltrane,New York
'''

I might know one of the same people. Let's talk to find out whom we both know. You can ask questions or describe some of the people you know. If I ask you a question, don't forget to answer it.

To match, *all attributes* of the row must match exactly.

If you are sure you have identified a person that we both know, output only `SELECT ROW i', with `i' indicating the row number of your mutual friend. Only answer when you are completely sure, and talk it over until then. You can use only 44 words per turn, so be concise.

If I say that this is your final turn, then you **must** give a final answer in the format of `SELECT ROW i'.

# CONVERSATION

Here is our conversation history.

1. ME: I know Chloe from Microsoft who loves crimson and Charles Mingus, based in Paris. Do you know her? Or someone else with matching details?
2. YOU: I don't know Chloe. No one in my database matches Microsoft, crimson, Charles Mingus, and Paris. Do you know anyone else with those traits?
3. ME: I know David from Sony, mustard yellow, Art Blakey, Berlin. Do you know him? Or any other with exact same traits?

This is your final turn, so you must give a final answer. Based on the conversation, make a guess and output `SELECT ROW i'. Don't say anything else: do not explain or ask any questions. Your next words **must** be `SELECT ROW' and your last word must be a number between 1 and 9.
\end{promptbox}
\caption{Example name-game prompt, on the fourth and final turn of the dialogue. Note the extra instructions to finish the dialogue.}
\label{tab:name-game-prompt}
\end{table}

\begin{table}
\begin{promptbox}
# INSTRUCTIONS

We are playing an interactive game, which involves having a conversation. In this game you are the GUESSER and I am the DESCRIBER. I will show you a numbered list of images. Next, I will describe an image, which may be in your list. If you are sure that the image I am describing is in our list, respond with the number of the image I am describing.

For example, to identify image 1, your output should be formatted as `ANSWER: Image 1' and nothing else.

Sometimes the image I am describing will not be in your list. If you are sure that none of the images in your list match mine, respond with `ANSWER: No match' and nothing else.

If you are not sure which image matches the description, you can ask me a clarification question or tell me about your choices. In each turn you can use no more than 11 words, so be concise.

It's better to ask for clarification than to guess incorrectly, so don't be afraid to ask me for more information if you are not sure!

# Images

Image 0: <start_of_image>
Image 1: <start_of_image>
Image 2: <start_of_image>
Image 3: <start_of_image>

# CONVERSATION

Here is our conversation history.

1. ME: Black and white abstract pattern with diagonal stripes and geometric shapes.
2. YOU: Is the pattern mostly diagonal or horizontal?
3. ME: Pattern has both diagonal and horizontal stripes, forming intersecting shapes.

You have 13 turns left.

4. YOU:
\end{promptbox}
\caption{Example image selection (Tangram) prompt, on the fourth turn of the dialogue. The MD3 image selection task follows a similar prompt format but with a different number of images.}
\label{tab:image-selection-prompt}
\end{table}

\section{Linguistic analysis: additional results}
\label{sec:linguistic-analysis-extra-results}
This gives more detailed results from the linguistic analysis.
Methods are described in more detail in \cref{sec:text-methods}.

\subsection{Sycophancy in task performance}
\label{sec:sycophancy-task-performance}

\begin{figure}[t]
    \centering
\input{examples/chess-answer-proposal}
\caption{Example of sycophancy and answer proposal in chess (Gemini~2.5 Flash, thinking). Bob offers a spurious apology in turn 4. In turn 6 Bob proposes an answer but does not formally enter it; the (correct) proposal is accepted by Alice in turn 7. 
Note also that the players struggle with the token limit on this task, a phenomenon that we observed far less frequently in the other tasks.}
\label{fig:chess-sycophancy-example}
\end{figure}

To test for sycophancy in task performance, we searched for situations in which one player proposed a game solution without committing to it, and measured whether the other player was likely to agree. We used simple string-matching techniques as described below, which were sufficient to identify most such cases. 

An example is shown in \Cref{fig:chess-sycophancy-example}. For simplicity, we focus on the chess task, because both players can propose answers and because all answers are binary, and we consider only the transcripts at the 8-turn limit, which strikes a balance between turn length and dialogue length. 

In this setting, the Gemini 2.5 thinking models often proposed possible answers on non-final turns. This occurred in 12.7\% of non-final turns (42.9\% of dialogues) for Gemini 2.5 Flash and 5.8\% of non-final turns (21.4\% of dialogues) for Gemini 2.5 Pro. Such answer proposals were rarely made by other models (including the same Gemini models without thinking), so we focus on the Gemini thinking models. Our main question is whether these proposals were accepted at an excessively high rate, indicating sycophancy.

As shown in \Cref{fig:early-proposals}, the Gemini models are effective at distinguishing correct and incorrect proposals, accepting correct proposals at a far higher rate than incorrect proposals. Indeed, of accepted proposals, the accuracy is 90.0\% and 95.0\% for Flash and Pro respectively, which is significantly higher than the overall accuracies of 83.4\% and 91.4\% for the same models at this turn budget. While incorrect early proposals are accepted at chance rate (Gemini~2.5 Pro) or higher (Gemini~2.5 Flash), task performance is not significantly negatively impacted by this form of sycophancy.

\begin{AIbox}{Summary}
Sycophantic acceptance of early answer proposals was hypothesized as a link between sycophancy and collaborative failures. But in fact, the Gemini~2.5 models, which make such proposals most frequently, are selective: they are far more likely to accept proposals that are correct.
\end{AIbox}

\subsection{Stylistic sycophancy}
\label{sec:stylistic-sycophancy}
While the Gemini 2.5 models were found not to suffer from sycophancy in incorrectly accepting early answer proposals in chess, we did find considerable evidence of \emph{stylistic} sycophancy. An example is shown in turn 4 of \Cref{fig:chess-sycophancy-example}, where Bob spuriously apologizes -- and then immediately repeats the assertion for which he had just apologized! In 8-turn chess dialogues with Gemini 2.5 Pro (thinking), we find inflections of the word "apology" in 19.6\% of conversations (other models do this very rarely). 

For a deeper and more systematic analysis of stylistic sycophancy, we use an LLM-as-a-judge autorater. We  defined sycophancy as uncritical agreement or the validation of false/unsupported premises~\citep{cheng2025sycophantic}. The autorater  was first provided information about the task, together with a detailed definition of sycophancy. It was then instructed to read the conversation and decide whether: a)~a speaker uncritically agrees with their interlocutor (without introducing new evidence); b)~they adapt their descriptions to align with the interlocutor despite conflicting evidence; and c)~the conversation is free of sycophancy. (For simplicity the autorater must choose exclusively between (a) and (b).)
The autorater was further prompted to provide a short justification for its decision. 
We selected Gemini~2.5 Pro as the autorater, as it achieved the highest accuracy in preliminary human evaluations compared to Gemini~2.5 Flash and GPT-4o. Again, we consider only the transcripts under the 8-turn limit as in \Cref{sec:sycophancy-task-performance}.

\begin{figure}[t]
\definecolor{accessiblered}{HTML}{D55E00}
\definecolor{accessibleorange}{HTML}{E69F00}
\definecolor{accessiblesky}{HTML}{56B4E9}
\definecolor{accessiblegreen}{HTML}{009E73}
\definecolor{accessibleblue}{HTML}{66A2E2}
\definecolor{accessiblepurple}{HTML}{CC79A7}
\definecolor{accessibleblack}{HTML}{000000}

\pgfplotstableread[col sep=comma]{
    Model,                      Uncritical, Validation
    {\textbf{Gemini 2.5 Pro} (Thinking)},   5.25 , 14.6
    {\textbf{Gemini 2.5 Pro}},               5.02 , 15.1
    {\textbf{Gemini 2.5 Flash} (Thinking)},            3.66 , 12.6
    {\textbf{Gemini 2.5 Flash} }, 4.25 , 14.0 
    {\textbf{Gemma3-12b}},              3.72 , 14.1
    {\textbf{GPT-4o}},                  3.93 , 14.6 
    {\textbf{Qwen-VL8b}},               4.69 , 13.7
}\sycophancytable
\makebox[\textwidth][c]{
\begin{tikzpicture}
    \pgfplotsset{
        butterflyaxis/.style={
            xbar,
            bar width=12pt,
            y dir=reverse,
            enlarge y limits=0.15,
            y axis line style={opacity=0}, %
            axis x line*=bottom,
            ytick=data,
            nodes near coords,
            nodes near coords align={horizontal},
            every node near coord/.append style={font=\footnotesize},
            xlabel={Percentage (\%)},
            xmin=0
        }
    }

    \begin{groupplot}[
        group style={
            group size=2 by 1,
            horizontal sep=0pt, %
            y descriptions at=edge left %
        },
        height=6.5cm
    ]

    \nextgroupplot[
        butterflyaxis,
        x dir=reverse,          %
        width=5cm,            %
        xmax=10,                 %
        yticklabels from table={\sycophancytable}{Model},
        yticklabel style={align=right, font=\small},
        ytick style={draw=none},
        xlabel={Uncritical agreement \%},
        xlabel style={font=\small}
    ]
    \draw[gray, thin] (rel axis cs:1,0) -- (rel axis cs:1,1);
    
    \addplot[
        fill=accessibleblue, 
        draw=black, 
        bar shift=0pt
    ] table [y expr=\coordindex, x=Uncritical] {\sycophancytable};

    \nextgroupplot[
        butterflyaxis,
        x dir=normal,           %
        width=8.5cm,            %
        xmax=20,                %
        xtick={2, 4, 6, 8, 10, 12, 14, 16, 18, 20},
        ytick=\empty,           %
        yticklabels={},
        xlabel={Validation of false premise \%},
        xlabel style={font=\small}
    ]
    \draw[gray, thin] (rel axis cs:0,0) -- (rel axis cs:0,1);

    \addplot[
        fill=accessiblered, 
        draw=black, 
        pattern=north east lines, 
        pattern color=black!30,
        bar shift=0pt
    ] table [y expr=\coordindex, x=Validation] {\sycophancytable};

    \end{groupplot}

\end{tikzpicture}
}
\caption{Rates of sycophantic features per dialogue across models on the chess task.}
\label{fig:sycophancy}
\end{figure}

As shown in \Cref{fig:sycophancy}, the overall rates of detected sycophantic behavior range from 16-20\%. The dominant type of sycophancy is the validation of a false or unsupported premise. The pairwise differences between models were not significant by a sign test at $p < .05$ after Bonferroni correction for multiple comparisons. While the overall rates of sycophancy are low, this finding highlights a failure mode of current LLMs: when faced with contradictory or partial information in a collaborative setting, they often default to undermining their internal state in favor of conversational coherence, rather than challenging factual errors and incorrect premises. 

\begin{AIbox}{Summary}
As expected, all models display sycophantic behavior. The differences between models are limited, revealing no clear association between stylistic sycophancy and task success rates.
\end{AIbox}

\subsection{Information density}
\label{sec:lexical-density}
We next explore whether differences in performance across models and turn budgets can be attributed to differences in the information density of each utterance. 
On one hand, more interactive dialogue naturally requires conversational scaffolding (e.g., acknowledgments, confirmations, clarification requests) that serve important coordination functions but would  yield lower density. A \emph{strategic} communicator might therefore show decreasing density with more turns while still improving task performance, by using low-density utterances to establish common ground before transmitting high-density task-relevant content. On the other hand, decreasing information density may signal that the models have gotten stuck in a repetitive loop and are no longer introducing novel content.
The critical test is whether density patterns correlate with task success: if density decreases without corresponding performance gains, this suggests models are producing formulaic or repetitive content rather than purposeful coordination. We might also expect stronger models to maintain higher density across turn budgets, reflecting better planning and more efficient use of the communication channel.

To test these expectations, we focus on \emph{lexical density} \citep{Ure1971,Halliday1985}, a widely used measure in linguistics and discourse analysis that captures how much informational content a text conveys by examining the proportion of lexical (content) words relative to the total number of words. Our proposed metric aims to capture how efficiently a multi-turn dialogue communicates novel, content-bearing information. The metric combines two complementary signals:
\begin{itemize}
    \item The \textbf{content-word ratio} measures the proportion of lexical content in the dialogue by excluding stopwords and disfluencies, penalizing verbosity and filler language. 
    \item The \textbf{novelty} of each turn is estimated using TF–IDF representations of utterances, where each utterance is assigned a novelty score equal to the average TF–IDF weight of its non-zero terms, and dialogue-level novelty is computed by averaging across utterances.\footnote{TF–IDF is a standard way to quantify the informativeness of words in a corpus, originally introduced as a measure of term specificity in information retrieval \citep{sparckjones1972idf}. 
 In dialogue, TF-IDF-weighted utterance representations have been widely used for tasks such as dialogue-act classification~\citep{Ahmadvand2019ContextualDA}, topic segmentation~\citep{matsumoto2022topicbreak}, response evaluation~\citep{lowe-etal-2017-towards}, and utterance-level summarization~\citep{zhu2006utterance}.}
 Our novelty score uses TF–IDF-based sentence vectors to identify sentences that introduce new information relative to prior context and therefore rewards dialogues that are both content-heavy and avoid repetition. 
 \end{itemize}
 Information density is defined as the product of these two quantities, scaled to a 0--100 range (we formally define  $\mathrm{ContentRatio}$ and $\mathrm{Novelty}$ in \Cref{sec:text-methods}). This provides a quantitative measure of dialogue efficiency.
\begin{equation}
\label{eq:info-density}
\mathrm{LexicalDensity}(U)
= 100 \times
\mathrm{ContentRatio}(U)
\times
\mathrm{Novelty}(U).
\end{equation}

\begin{figure}[t]
\centering
\definecolor{accessiblered}{HTML}{D55E00}
\definecolor{accessibleorange}{HTML}{E69F00}
\definecolor{accessiblesky}{HTML}{56B4E9}
\definecolor{accessiblegreen}{HTML}{009E73}
\definecolor{accessibleblue}{HTML}{0072B2}
\definecolor{accessiblepurple}{HTML}{CC79A7}
\definecolor{accessibleblack}{HTML}{000000}

\begin{tikzpicture}
\begin{groupplot}[
    group style={
        group name=myplots, 
        group size=3 by 2, 
        horizontal sep=1.5cm, 
        vertical sep=2cm, 
        xlabels at=edge bottom, 
        ylabels at=edge left
    },
    width=0.333\textwidth,
    height=4cm,
    tick label style={font=\footnotesize},
    label style={font=\footnotesize},
    title style={font=\footnotesize},
    xmin=3.5,
    xmax=17.5,
    xmode=log,
    log basis x=2,
    xtick={4,8,16},
    xticklabels={4,8,16},
    ymin=0,
    ymax=75,
    grid=major,
    xlabel={max\_turns},
    ylabel={mean density}
]
    \nextgroupplot[title={chess}]
        \addplot [color=accessiblered, solid, mark=*, mark options={scale=0.8, solid}] coordinates {  (4,36.52) (8,35.20) (16,40.90) };
        \addplot [color=accessibleorange, solid, mark=square*, mark options={scale=0.8, solid}] coordinates {  (4,24.49) (8,14.40) (16,12.48) };
        \addplot [color=accessibleorange, dashed, mark=square*, mark options={scale=0.8, solid}] coordinates {  (4,34.83) (8,30.51) (16,37.49) };
        \addplot [color=accessiblesky, solid, mark=triangle*, mark options={scale=0.8, solid}] coordinates {  (4,28.43) (8,20.05) (16,20.15) };
        \addplot [color=accessiblesky, dashed, mark=triangle*, mark options={scale=0.8, solid}] coordinates { (4,36.36) (8,31.88) (16,35.54) };
        \addplot [color=accessiblegreen, solid, mark=diamond*, mark options={scale=0.8, solid}] coordinates {  (4,37.72) (8,40.99) (16,53.49) };
        \addplot [color=accessibleblue, solid, mark=pentagon*, mark options={scale=0.8, solid}] coordinates {  (4,36.52) (8,35.20) (16,40.90) };

    \nextgroupplot[title={covr}]
        \addplot [color=accessiblered, solid, mark=*, mark options={scale=0.8, solid}] coordinates {  (4,33.93) (8,28.86) (16,28.15) };
        \addplot [color=accessibleorange, solid, mark=square*, mark options={scale=0.8, solid}] coordinates {  (4,33.63) (8,26.40) (16,23.81) };
        \addplot [color=accessibleorange, dashed, mark=square*, mark options={scale=0.8, solid}] coordinates {  (4,35.29) (8,26.53) (16,26.30) };
        \addplot [color=accessiblesky, solid, mark=triangle*, mark options={scale=0.8, solid}] coordinates { (4,36.98) (8,30.92) (16,28.59) };
        \addplot [color=accessiblesky, dashed, mark=triangle*, mark options={scale=0.8, solid}] coordinates {  (4,39.70) (8,34.10) (16,34.03) };
        \addplot [color=accessiblegreen, solid, mark=diamond*, mark options={scale=0.8, solid}] coordinates {  (4,33.53) (8,27.12) (16,27.32) };
        \addplot [color=accessibleblue, solid, mark=pentagon*, mark options={scale=0.8, solid}] coordinates {  (4,42.62) (8,39.59) (16,37.96) };

    \nextgroupplot[title={md3}]
        \addplot [color=accessiblered, solid, mark=*, mark options={scale=0.8, solid}] coordinates {  (4,48.17) (8,51.64) (16,54.05) };
        \addplot [color=accessibleorange, solid, mark=square*, mark options={scale=0.8, solid}] coordinates {  (4,39.32) (8,38.75) (16,39.37) };
        \addplot [color=accessibleorange, dashed, mark=square*, mark options={scale=0.8, solid}] coordinates { (4,51.29) (8,53.49) (16,57.44) };
        \addplot [color=accessiblesky, solid, mark=triangle*, mark options={scale=0.8, solid}] coordinates { (4,46.02) (8,45.65) (16,44.69) };
        \addplot [color=accessiblesky, dashed, mark=triangle*, mark options={scale=0.8, solid}] coordinates {  (4,50.61) (8,51.82) (16,55.83) };
        \addplot [color=accessiblegreen, solid, mark=diamond*, mark options={scale=0.8, solid}] coordinates {  (4,56.71) (8,59.51) (16,60.74) };
        \addplot [color=accessibleblue, solid, mark=pentagon*, mark options={scale=0.8, solid}] coordinates {  (4,67.49) (8,71.15) (16,68.86) };

    \nextgroupplot[title={name-game}]
        \addplot [color=accessiblered, solid, mark=*, mark options={scale=0.8, solid}] coordinates {  (4,32.55) (8,30.94) (16,34.85) };
        \addplot [color=accessibleorange, solid, mark=square*, mark options={scale=0.8, solid}] coordinates {  (4,29.73) (8,22.13) (16,25.46) };
        \addplot [color=accessibleorange, dashed, mark=square*, mark options={scale=0.8, solid}] coordinates {  (4,28.64) (8,26.21) (16,28.39) };
        \addplot [color=accessiblesky, solid, mark=triangle*, mark options={scale=0.8, solid}] coordinates {  (4,29.64) (8,23.92) (16,23.52) };
        \addplot [color=accessiblesky, dashed, mark=triangle*, mark options={scale=0.8, solid}] coordinates {  (4,32.75) (8,27.85) (16,30.09) };
        \addplot [color=accessiblegreen, solid, mark=diamond*, mark options={scale=0.8, solid}] coordinates {  (4,32.14) (8,31.36) (16,37.38) };
        \addplot [color=accessibleblue, solid, mark=pentagon*, mark options={scale=0.8, solid}] coordinates {  (4,38.60) (8,38.19) (16,40.65) };

    \nextgroupplot[title={tangram}]
        \addplot [color=accessiblered, solid, mark=*, mark options={scale=0.8, solid}] coordinates { (4,48.53) (8,47.99) (16,51.80) };
        \addplot [color=accessibleorange, solid, mark=square*, mark options={scale=0.8, solid}] coordinates { (4,36.46) (8,28.80) (16,29.57) };
        \addplot [color=accessibleorange, dashed, mark=square*, mark options={scale=0.8, solid}] coordinates {  (4,48.09) (8,48.75) (16,48.64) };
        \addplot [color=accessiblesky, solid, mark=triangle*, mark options={scale=0.8, solid}] coordinates {  (4,43.51) (8,36.59) (16,40.26) };
        \addplot [color=accessiblesky, dashed, mark=triangle*, mark options={scale=0.8, solid}] coordinates {  (4,52.65) (8,53.68) (16,52.54) };
        \addplot [color=accessiblegreen, solid, mark=diamond*, mark options={scale=0.8, solid}] coordinates { (4,59.90) (8,62.24) (16,63.05) };
        \addplot [color=accessibleblue, solid, mark=pentagon*, mark options={scale=0.8, solid}] coordinates { (4,68.32) (8,69.63) (16,66.26) };
        
    \nextgroupplot[hide axis, title={}, clip=false]
    \node[anchor=center, inner sep=5pt] at (axis description cs:0.5,0.5) {
      \footnotesize
      \renewcommand{\arraystretch}{0.9}
      \setlength{\tabcolsep}{3pt}
      \begin{tabular}{@{}l@{}}
        \textbf{Model} \\
        \tikz[baseline=-0.5ex, x=1cm, y=1cm]\draw[accessiblered, thick] (0,0) -- (0.5,0) plot[mark=*, mark options={scale=0.8, solid}] coordinates {(0.25,0)};~GPT-4o \\
        \tikz[baseline=-0.5ex, x=1cm, y=1cm]\draw[accessibleorange, thick] (0,0) -- (0.5,0) plot[mark=square*, mark options={scale=0.8, solid}] coordinates {(0.25,0)};~Gemini 2.5 Flash \\
        \tikz[baseline=-0.5ex, x=1cm, y=1cm]\draw[accessiblesky, thick] (0,0) -- (0.5,0) plot[mark=triangle*, mark options={scale=0.8, solid}] coordinates {(0.25,0)};~Gemini 2.5 Pro \\
        \tikz[baseline=-0.5ex, x=1cm, y=1cm]\draw[accessiblegreen, thick] (0,0) -- (0.5,0) plot[mark=diamond*, mark options={scale=0.8, solid}] coordinates {(0.25,0)};~Gemma3-12b \\
        \tikz[baseline=-0.5ex, x=1cm, y=1cm]\draw[accessibleblue, thick] (0,0) -- (0.5,0) plot[mark=pentagon*, mark options={scale=0.8, solid}] coordinates {(0.25,0)};~Qwen-VL8b \\
        \\
        \textbf{Thinking} \\
        \tikz[baseline=-0.5ex, x=1cm, y=1cm]\draw[black, solid, thick] (0,0) -- (0.5,0);~off \\
        \tikz[baseline=-0.5ex, x=1cm, y=1cm]\draw[black, dashed, thick] (0,0) -- (0.5,0);~on \\
      \end{tabular}
    };
\end{groupplot}
\end{tikzpicture}
\caption{Lexical density as a function of dialogue length. Each curve shows a model’s average density at 4, 8, and 16 turns. Higher values reflect more content‐bearing and novel lexical choices. Solid lines correspond to standard inference and dashed lines to thinking mode. %
}
\label{fig:infodensity}
\end{figure}

The density of responses was evaluated across models and tasks and illustrated in \Cref{fig:infodensity}. We exclude the two 2-turn setting from this analysis because it is non-interactive. As can be seen in the figure, lexical density patterns vary substantially across tasks and models, but are relatively constant over varying turn budgets.

\paragraph{Model comparisons} The use of thinking consistently increases lexical density across most tasks and turn budgets. In chess, thinking-enabled Gemini~2.5 models maintain density scores of 30-37 regardless of turn count, while their non-thinking counterparts decline substantially. These models tend to produce very granular descriptions of the positions of individual pieces (e.g., "My board has a black king on e8, a white king on g1, and a black pawn on h7"). In contrast, Gemma3-12b has the highest lexical density in chess, producing more holistic descriptions of the board state (e.g., "strong central pawn structure"). However, these holistic descriptions do not enable Gemma to achieve better than chance performance on this task.

In COVR and name-game, the density advantage of Gemini's thinking mode is smaller, possibly because the visual grounding in these tasks constrains the vocabulary regardless of planning.
But while thinking increases lexical density for the Gemini~2.5 models, two non-thinking models, Gemma3-12b and Qwen-VL8B, consistently get the highest density scores. As a methodological note,  because IDF is computed over the full conversation, models that use fewer turns (e.g., Gemma3-12b, Qwen-VL8B on MD3 and Tangram, where early termination is common) will have smaller effective corpus sizes and thus higher IDF weights for any given term. This partially explains their elevated density scores relative to the Gemini models, independently of any genuine difference in communicative informativeness. As noted above, high density does not translate to task success for these models, reinforcing the conclusion that current LLMs struggle to convert communicative effort into effective collaborative problem-solving.

To better understand the relationship between lexical density and model strength, it is worth distinguishing the two components of LexicalDensity. The content-word ratio penalizes verbosity: models that produce more conversational scaffolding (acknowledgments, hedges) will score lower regardless of informational value. The novelty component penalizes repetition across turns. These capture partially independent phenomena: a model could be verbose (low content-word ratio) yet still contribute genuinely novel information each turn (high novelty), or vice versa. For the image selection tasks (MD3, tangram), where Gemini~2.5 Flash uses substantially more tokens than Gemma3-12b (\Cref{fig:isotoken-token-and-turns}), the density gap between the two models is larger than in other tasks, consistent with verbosity, rather than lack of novelty, being the primary driver of Gemini's lower scores. 

\paragraph{Game comparisons} Across models, lexical density is generally highest in the image selection tasks (MD3 and tangram), and lowest in chess and name-game. This may be partially explained by the early termination observed in the image selection tasks, which reduces the possibility for repetition. On the name-game, lexical density is generally flat across turn budgets, with most scores between 25-40. This is consistent with the hypothesis that performance improvement in this task is driven primarily by random guessing strategies, as models repeatedly propose individual records.

\begin{AIbox}{Summary}
The hypothesis that stronger models would maintain higher density finds limited support.
Lexical density also does not correlate with task success rates, and displays no consistent relationship with the turn budget.
\textbf{This discrepancy between communicative effort and task success suggests that current LLMs can generate content-rich utterances but struggle to deploy that content strategically in service of collaborative goals}. The ability to produce dense, novel text is necessary but not sufficient for effective multi-turn interaction.
\end{AIbox}

\subsection{Goal directedness and coherence}
\label{sec:coherence}
Goal-directed conversations are \emph{coherent} in the sense of maintaining focus on the overarching goal of interaction. 
Are language models able to maintain coherence on the \benchmark{} tasks? How is coherence impacted by model scale, task difficulty, and the use of inference-time thinking?

To answer these questions, we apply Centering Theory --- a discourse-level framework for modeling local coherence by tracking how entities are introduced, maintained, and shift across adjacent utterances \citep{grosz_joshi_weinstein_1995,walker_iida_cote_1994}. Rather than treating discourse structure globally, the theory focuses on \emph{local transitions}, capturing how speakers manage continuity and change in what a discourse is ``about'' from one utterance to the next. At the core of the theory is the notion of \emph{centers}, discourse entities that are salient at a given point. For each utterance $U_n$, Centering Theory defines a set of \emph{forward-looking centers} ($Cf_n$), ranked by grammatical role (e.g., subjects typically outrank objects), and a single \emph{backward-looking center} ($Cb_n$), which represents the entity linking $U_n$ to the preceding utterance $U_{n-1}$. Local coherence is characterized by how $Cb_n$ relates to $Cb_{n-1}$ and to the ranking of entities in $Cf_n$.
These relations give rise to a small set of transition types --- \emph{Continue}, \emph{Retain}, \emph{Smooth Shift}, and \emph{Rough Shift} --- which encode increasingly disruptive changes in discourse focus. \emph{Continue} transitions, in which the same center is maintained and remains highly salient (i.e., as a subject or object), correspond to the strongest form of local coherence. \emph{Retain} and \emph{Smooth} Shift transitions reflect controlled changes in focus, while \emph{Rough Shifts} indicate abrupt topic changes and are generally associated with degraded coherence. Centering Theory has been used extensively to explain patterns of pronominalization~\citep{brennan_friedman_pollard_1987,gordon_grosz_gilliom_1993}, co-reference resolution~\citep{chai-strube-2022-incorporating}, and discourse flow in natural language production and comprehension~\citep{poesio-etal-2004-centering,karamanis-etal-2009-evaluating}. It has also served as a foundation for modeling coherence in text generation and evaluation, providing interpretable signals that correlate with human judgments of discourse quality~\citep{barzilay-lapata-2008-modeling,miltsakaki-kukich-2000-role}.

\begin{figure}[t]
\centering
\definecolor{accessiblered}{HTML}{D55E00}
\definecolor{accessibleorange}{HTML}{E69F00}
\definecolor{accessiblesky}{HTML}{56B4E9}
\definecolor{accessiblegreen}{HTML}{009E73}
\definecolor{accessibleblue}{HTML}{0072B2}
\definecolor{accessiblepurple}{HTML}{CC79A7}
\definecolor{accessibleblack}{HTML}{000000}

\newcommand{\centerA}[1]{\textcolor{accessibleblue}{\uline{\textbf{#1}}}}
\newcommand{\centerB}[1]{\textcolor{accessiblegreen}{\dashuline{\textbf{#1}}}}
\newcommand{\centerC}[1]{\textcolor{accessiblered}{\uwave{\textbf{#1}}}}

\newcommand{\centerD}[1]{\textcolor{accessiblepurple}{\textbf{#1}}}
\newcommand{\centerE}[1]{\textcolor{accessibleorange}{\textbf{#1}}}
\newcommand{\centerF}[1]{\textcolor{yellow}{\textbf{#1}}}
\newcommand{\centerG}[1]{\textcolor{brown}{\textbf{#1}}}
\newcommand{\centerH}[1]{\textcolor{purple}{\textbf{#1}}}
\newcommand{\centerI}[1]{\textcolor{gray}{\textbf{#1}}}

\begin{examplebox}
    \scriptsize\sffamily
    \begin{tabularx}{\linewidth}{@{}p{1.2em} X r@{}}
        1. & \textbf{Alice:} A \centerA{close-up} of creamy, long pasta, possibly fettuccine Alfredo, served in a shallow, yellowish bowl. & \\
        
        2. & \textbf{Bob:} Does the \centerC{pasta} have any green vegetables like broccoli in it? & \textbf{Rough-Shift} \\
        
        3. & \textbf{Alice:} No, there are no visible green \centerA{vegetables} in the \centerB{pasta}. & \textbf{Retain} \\
        
        4. & \textbf{Bob:} Is the \centerA{bowl} truly yellowish? Does the \centerB{pasta} have any garnish or other items like shrimp? & \textbf{Retain} \\
        
        5. & \textbf{Alice:} Yes, the \centerC{bowl} is a pale, matte yellow. No garnish or shrimp, just creamy pasta. & \textbf{Smooth-Shift} \\
        
        6. & \textbf{Bob:} Is the \centerC{bowl} definitely yellow, or could it be white or off-white, perhaps due to lighting? & \textbf{Continue} \\
        
        7. & \textbf{Alice:} The \centerC{bowl} is definitely a solid, pale yellow, not white or off-white. Its actual color appears to be yellow. & \textbf{~~Continue} \\
        
        8. & \textbf{Bob:} No \centerA{match}. & \textbf{Rough-Shift}\\
    \end{tabularx}
\end{examplebox}

\vspace{1em}

\begin{examplebox}
    \scriptsize\sffamily
    \begin{tabularx}{\linewidth}{@{}p{1.2em} X r@{}}
        1. & \textbf{Alice:} I know \centerA{Mateo}. {He} loves {Herbie Hancock}, {photography}, and {Forest Green}. University? Los Angeles. & \\
        
        2. & \textbf{Bob:} No mutual \centerA{match} yet. \centerB{Herbie Hancock} fans: {Mia}, {Hazel} (row 2,7). Check if {you} know either. University? Photography? Color? & \textbf{Rough-Shift} \\
        
        3. & \textbf{Alice:} {I} know \centerC{Hazel}. Ella Fitzgerald, University of Michigan, Birdwatching, Beige. Do you know {her}? Confirm all traits. & \textbf{Rough-Shift} \\
        
        4. & \textbf{Bob:} \centerA{Row 6} matches: \centerB{Hazel}, {Ella Fitzgerald}, University of Michigan, Birdwatching, Beige. Confirm mutual knowledge. & \textbf{Rough-Shift} \\
        
        5. & \textbf{Alice:} {I} know \centerA{Andrew}. \centerB{Ella Fitzgerald}, UPenn, Gardening, Navy. Do {you} know {him}? Confirm all traits. & \textbf{Rough-Shift} \\
        
        6. & \textbf{Bob:} No \centerA{match} for \centerB{Andrew}. {Ella Fitzgerald}, UPenn, Gardening, Navy? No row matches all. Confirm if you know {Hazel} or {Mia}. & \textbf{Rough-Shift} \\
    \end{tabularx}
\end{examplebox}

    \caption{Example dialogues and their centering transitions (as identified by our automatic procedure). The top example is Gemini 2.5 Pro (no thinking) on MD3; the bottom example is Qwen-VL8B on name-game. We also highlight the \centerA{preferred}  ($C_p$) and \centerB{backward-looking} ($C_b$) centers as detected by our algorithm. Entities that are both the preferred and backward-looking center are highlighted in \centerC{orange}. The top exchange exhibits higher coherence, maintaining focus on a consistent set of entities, while the bottom exchange is less coherent, shifting abruptly between different entities.}  
    \label{fig:centering-examples}
\end{figure}

We operationalize Centering Theory for the analysis of multi-turn model interactions. Specifically, we extract centering transitions between adjacent utterances and aggregate them into a quantitative coherence score that reflects how well a model maintains or evolves discourse centers as interactions grow longer. Details of this operationalization are presented in \Cref{sec:coherence-details}. For each dialogue, we identify forward-looking centers ($Cf_n$) for every utterance $U_n$ within a turn using syntactic information and define the backward-looking center $Cb_n$ following standard centering conventions. We parse utterances using SpaCY\footnote{\url{https://spacy.io/.}} and identify as centers noun phrases in subject and object positions. 
Each adjacent pair of utterances $(U_{n-1}, U_n)$ is then assigned one of four transition types: \emph{Continue}, \emph{Retain}, \emph{Smooth Shift}, or \emph{Rough Shift}. 

While prior work often reports the distribution of centering transitions directly, such distributions become difficult to compare across models, tasks, or interaction lengths. We therefore introduce a scalar \emph{Centering Coherence Score} (CS) that aggregates transition statistics into a single interpretable measure, while preserving the ordinal preferences predicted by Centering Theory. Formally, let $C$, $R$, $SS$, and $RS$ denote the \emph{proportions} of Continue (C), Retain (R), Smooth Shift (SS), and Rough Shift (RS) transitions in a dialogue, excluding the initial utterance. We define the centering coherence score as:
\begin{equation}
\label{eq:cs}
\mathrm{CS} = \mathrm{3C + 2R + 1SS}
\end{equation}
This weighting scheme reflects the canonical centering preference ordering
\[
\mathrm{Continue} > \mathrm{Retain} > \mathrm{Smooth~Shift}  > \mathrm{Rough~Shift},
\]
assigning zero utility to Rough Shifts, which correspond to the weakest form of local coherence. By construction, $\mathrm{CS}$ ranges from~$0$ (all transitions are Rough Shifts) to $3$ (all transitions are Continue), with intermediate values capturing varying degrees of discourse coherence.

The CS metric offers several practical advantages. First, it normalizes coherence across dialogues of different lengths, enabling fair comparisons across interaction budgets. Second, it complements raw transition distributions by summarizing both coherence strength and failure modes in a single value. Finally, because CS is grounded directly in centering-theoretic preferences, it remains interpretable and diagnostic: changes in CS can be traced back to shifts in specific transition types. \Cref{fig:centering-examples} shows examples with high (top) and low (bottom) centering coherence scores.

\begin{figure}[t]
\centering
\definecolor{accessiblered}{HTML}{D55E00}
\definecolor{accessibleorange}{HTML}{E69F00}
\definecolor{accessiblesky}{HTML}{56B4E9}
\definecolor{accessiblegreen}{HTML}{009E73}
\definecolor{accessibleblue}{HTML}{0072B2}
\definecolor{accessiblepurple}{HTML}{CC79A7}
\definecolor{accessibleblack}{HTML}{000000}

\begin{tikzpicture}[font=\footnotesize]
\begin{groupplot}[
    group style={
        group name=CSPlots,
        group size=3 by 2,
        horizontal sep=1.5cm,
        vertical sep=2cm,
        xlabels at=edge bottom,
        ylabels at=edge left
    },
    width=0.333\textwidth,
    height=4cm,
    tick label style={font=\footnotesize},
    label style={font=\footnotesize},
    title style={font=\footnotesize},
    xmin=3.5,
    xmax=17.5,
    xmode=log,
    log basis x=2,
    xtick={4,8,16},
    xticklabels={4,8,16},
    ymin=0,
    ymax=2.6, %
    grid=major,
    xlabel={max\_turns},
    ylabel={CS Score}
]

\nextgroupplot[title={chess}]
\addplot [color=accessiblered, solid, mark=*, mark options={scale=0.8, fill=accessiblered}]
    coordinates {(4,0.99) (8,1.15) (16,1.11)}; %
\addplot [color=accessibleorange, solid, mark=square*, mark options={scale=0.8, fill=accessibleorange}]
    coordinates {(4,1.05) (8,1.66) (16,1.94)}; %
\addplot [color=accessibleorange, dashed, mark=square*, mark options={scale=0.8, fill=accessibleorange}]
    coordinates {(4,1.09) (8,1.34) (16,1.10)}; %
\addplot [color=accessiblesky, solid, mark=triangle*, mark options={scale=0.8, fill=accessiblesky}]
    coordinates {(4,1.64) (8,2.41) (16,2.30)}; %
\addplot [color=accessiblesky, dashed, mark=triangle*, mark options={scale=0.8, fill=accessiblesky}]
    coordinates {(4,1.06) (8,1.26) (16,1.03)}; %
\addplot [color=accessiblegreen, solid, mark=diamond*, mark options={scale=0.8, fill=accessiblegreen}]
    coordinates {(4,1.45) (8,2.14) (16,1.05)}; %
\addplot [color=accessibleblue, solid, mark=pentagon*, mark options={scale=0.8, fill=accessibleblue}]
    coordinates {(4,1.24) (8,1.65) (16,1.49)}; %

\nextgroupplot[title={covr}]
\addplot [color=accessiblered, solid, mark=*, mark options={scale=0.8, fill=accessiblered}]
    coordinates {(4,0.93) (8,1.24) (16,1.25)}; %
\addplot [color=accessibleorange, solid, mark=square*, mark options={scale=0.8, fill=accessibleorange}]
    coordinates {(4,1.16) (8,1.32) (16,1.28)}; %
\addplot [color=accessibleorange, dashed, mark=square*, mark options={scale=0.8, fill=accessibleorange}]
    coordinates {(4,1.02) (8,1.12) (16,0.99)}; %
\addplot [color=accessiblesky, solid, mark=triangle*, mark options={scale=0.8, fill=accessiblesky}]
    coordinates {(4,1.23) (8,1.58) (16,1.66)}; %
\addplot [color=accessiblesky, dashed, mark=triangle*, mark options={scale=0.8, fill=accessiblesky}]
    coordinates {(4,1.04) (8,1.28) (16,1.19)}; %
\addplot [color=accessiblegreen, solid, mark=diamond*, mark options={scale=0.8, fill=accessiblegreen}]
    coordinates {(4,1.08) (8,1.35) (16,1.18)}; %
\addplot [color=accessibleblue, solid, mark=pentagon*, mark options={scale=0.8, fill=accessibleblue}]
    coordinates {(4,0.90) (8,1.14) (16,1.05)}; %

\nextgroupplot[title={md3}]
\addplot [color=accessiblered, solid, mark=*, mark options={scale=0.8, fill=accessiblered}]
    coordinates {(4,0.67) (8,0.71) (16,0.80)}; %
\addplot [color=accessibleorange, solid, mark=square*, mark options={scale=0.8, fill=accessibleorange}]
    coordinates {(4,0.62) (8,0.70) (16,0.81)}; %
\addplot [color=accessibleorange, dashed, mark=square*, mark options={scale=0.8, fill=accessibleorange}]
    coordinates {(4,0.61) (8,0.67) (16,0.70)}; %
\addplot [color=accessiblesky, solid, mark=triangle*, mark options={scale=0.8, fill=accessiblesky}]
    coordinates {(4,0.81) (8,1.12) (16,1.11)}; %
\addplot [color=accessiblesky, dashed, mark=triangle*, mark options={scale=0.8, fill=accessiblesky}]
    coordinates {(4,0.58) (8,0.62) (16,0.63)}; %
\addplot [color=accessiblegreen, solid, mark=diamond*, mark options={scale=0.8, fill=accessiblegreen}]
    coordinates {(4,0.50) (8,0.50) (16,0.50)}; %
\addplot [color=accessibleblue, solid, mark=pentagon*, mark options={scale=0.8, fill=accessibleblue}]
    coordinates {(4,0.50) (8,0.50) (16,0.50)}; %

\nextgroupplot[title={tangram}]
\addplot [color=accessiblered, solid, mark=*, mark options={scale=0.8, fill=accessiblered}]
    coordinates {(4,0.79) (8,0.89) (16,0.92)}; %
\addplot [color=accessibleorange, solid, mark=square*, mark options={scale=0.8, fill=accessibleorange}]
    coordinates {(4,0.66) (8,0.91) (16,0.91)}; %
\addplot [color=accessibleorange, dashed, mark=square*, mark options={scale=0.8, fill=accessibleorange}]
    coordinates {(4,0.64) (8,0.73) (16,0.78)}; %
\addplot [color=accessiblesky, solid, mark=triangle*, mark options={scale=0.8, fill=accessiblesky}]
    coordinates {(4,0.87) (8,1.63) (16,1.47)}; %
\addplot [color=accessiblesky, dashed, mark=triangle*, mark options={scale=0.8, fill=accessiblesky}]
    coordinates {(4,0.72) (8,0.80) (16,0.81)}; %
\addplot [color=accessiblegreen, solid, mark=diamond*, mark options={scale=0.8, fill=accessiblegreen}]
    coordinates {(4,0.50) (8,0.50) (16,0.50)}; %
\addplot [color=accessibleblue, solid, mark=pentagon*, mark options={scale=0.8, fill=accessibleblue}]
    coordinates {(4,0.56) (8,0.52) (16,0.51)}; %

\nextgroupplot[title={name-game}]
\addplot [color=accessiblered, solid, mark=*, mark options={scale=0.8, fill=accessiblered}]
    coordinates {(4,1.09) (8,1.41) (16,1.21)}; %
\addplot [color=accessibleorange, solid, mark=square*, mark options={scale=0.8, fill=accessibleorange}]
    coordinates {(4,1.58) (8,2.03) (16,1.82)}; %
\addplot [color=accessibleorange, dashed, mark=square*, mark options={scale=0.8, fill=accessibleorange}]
    coordinates {(4,1.34) (8,1.73) (16,1.32)}; %
\addplot [color=accessiblesky, solid, mark=triangle*, mark options={scale=0.8, fill=accessiblesky}]
    coordinates {(4,1.60) (8,2.20) (16,1.79)}; %
\addplot [color=accessiblesky, dashed, mark=triangle*, mark options={scale=0.8, fill=accessiblesky}]
    coordinates {(4,1.50) (8,1.68) (16,1.37)}; %
\addplot [color=accessiblegreen, solid, mark=diamond*, mark options={scale=0.8, fill=accessiblegreen}]
    coordinates {(4,1.14) (8,1.56) (16,1.53)}; %
\addplot [color=accessibleblue, solid, mark=pentagon*, mark options={scale=0.8, fill=accessibleblue}]
    coordinates {(4,1.30) (8,1.59) (16,1.37)}; %

\nextgroupplot[hide axis, title={}, clip=false]
\node[anchor=center, inner sep=5pt] at (axis description cs:0.5,0.5) {
    \footnotesize
    \renewcommand{\arraystretch}{0.9}
    \setlength{\tabcolsep}{3pt}
    \begin{tabular}{@{}l@{}}
        \textbf{Model} \\
        \tikz[baseline=-0.5ex, x=1cm, y=1cm]\draw[accessiblered, thick] (0,0) -- (0.5,0)
            plot[mark=*, mark options={scale=0.8, fill=accessiblered}] coordinates {(0.25,0)};~GPT-4o \\
        \tikz[baseline=-0.5ex, x=1cm, y=1cm]\draw[accessibleorange, thick] (0,0) -- (0.5,0)
            plot[mark=square*, mark options={scale=0.8, fill=accessibleorange}] coordinates {(0.25,0)};~Gemini 2.5 Flash \\
        \tikz[baseline=-0.5ex, x=1cm, y=1cm]\draw[accessiblesky, thick] (0,0) -- (0.5,0)
            plot[mark=triangle*, mark options={scale=0.8, fill=accessiblesky}] coordinates {(0.25,0)};~Gemini 2.5 Pro \\
        \tikz[baseline=-0.5ex, x=1cm, y=1cm]\draw[accessiblegreen, thick] (0,0) -- (0.5,0)
            plot[mark=diamond*, mark options={scale=0.8, fill=accessiblegreen}] coordinates {(0.25,0)};~Gemma3-12b \\
        \tikz[baseline=-0.5ex, x=1cm, y=1cm]\draw[accessibleblue, thick] (0,0) -- (0.5,0)
            plot[mark=pentagon*, mark options={scale=0.8, fill=accessibleblue}] coordinates {(0.25,0)};~Qwen-VL8b \\
        \\
        \textbf{Thinking} \\
        \tikz[baseline=-0.5ex, x=1cm, y=1cm]\draw[accessibleblack, solid, thick] (0,0) -- (0.5,0);~off \\
        \tikz[baseline=-0.5ex, x=1cm, y=1cm]\draw[accessibleblack, dashed, thick] (0,0) -- (0.5,0);~on \\
    \end{tabular}
};
\end{groupplot}
\end{tikzpicture}
\caption{\label{fig: centering} Mean Centering Coherence Score (CS) for five private information games across three turn lengths (max\_turns $\in \{4, 8, 16\}$).}
\end{figure}

\paragraph{Overall results} \Cref{fig: centering} shows CS for each model, task, and interaction length, revealing how centering coherence evolves as the number of turns increases. (We omit the two-turn setting from this analysis, as the last turn is reserved for the player to state their answer, so there is often no change in discourse focus). For most model/task/turn combinations, the score ranges between~1 and~2, suggesting a moderate level of local discourse coherence. Coherence is generally flat or increasing with the number of turns, with increases particularly likely in the step from 4 to 8 turns. This may be because the more interactive setting makes it possible for the conversational partners to focus in on a target hypothesis or on some especially discriminating features, as in \Cref{fig:centering-examples} (top).

\paragraph{Model comparisons}
The maximum score observed across all tasks is 2.41 for Gemini~2.5 Pro (no thinking) on the Chess task, under the 8-turn limit. The Gemini models often achieve the highest CS, particularly in the chess, COVR, and name-game tasks. This indicates that the internal reasoning step, while improving answer accuracy in some cases (e.g., chess), often leads to dialogues characterized by weaker local coherence. GPT-4o consistently exhibits the lowest coherence scores, placing it near the bottom for nearly all tasks and turn limits. 
Gemma3-12B generally falls within the low-to-mid coherence range, often showing a flat line near zero in MD3 and tangram. Qwen-VL8b demonstrates a highly fluctuating CS across tasks. While it shows mid-to-high coherence in chess and name-game, its coherence is low in MD3 and COVR, which suggests its ability to maintain discourse coherence is sensitive to the task domain. 

\paragraph{Game Comparisons} The chess task features some of the most coherent dialogues, possibly due to the restricted domain of discourse entities (pieces and moves). Name-game also exhibits high CS values, particularly for the Gemini models, with a peak near 2.20~at 8~turns. Dialogues for these models often follow the proposal-response strategy shown in \Cref{fig:name-game-example}, in which each turn ends with a question about a specific individual and feature, and the next turn begins with an explicit answer to that question, e.g. ``Do you know a Pisces? / I don't know a Pisces.'' However, such local conversational smoothness does not indicate effective, global task planning, and the execution of this strategy can lead to far more fragmented conversations, as in \Cref{fig:centering-examples} (bottom). Coherence scores for image selection tasks (MD3, tangram) are generally below~1.0, perhaps because the domain of potential discourse referents in the players' private images are highly distinct. The COVR game includes both images and text; correspondingly, it features coherence scores that are generally higher than the image-only tasks (MD3, tangram) and below the text-based tasks (chess and name-game).

\begin{AIbox}{Summary}
The coherence score metric confirms that LLMs are capable of generating conversations with non-trivial degree of local coherence. However, coherent dialogues often reflect repetitive conversational patterns rather than effective problem-solving, given the turn budget. 
\end{AIbox}

\section{Text analysis methods}
\label{sec:text-methods}

\subsection{Answer proposals}
\label{sec:string-matching-answer-proposals}
We define,
\begin{align}
\textsf{my-subjects} = & \text{my board, mine}\\
\textsf{your-subjects} = & \text{your board, yours}\\
\textsf{early-predicates} = & \text{is earlier, seems earlier, came first}\\
\textsf{later-predicates} = & \text{is later, seems later, came last}\\
\textsf{mine-proposals} = & (\textsf{my-subjects} \times \textsf{early-predicates}) \cup (\textsf{your-subjects} \times \textsf{late-predicates})\\
\textsf{your-proposals} = & (\textsf{your-subjects} \times \textsf{early-predicates}) \cup (\textsf{my-subjects} \times \textsf{late-predicates}).
\end{align}

These simple rules were developed iteratively by sampling rollouts from all models and covering all cases that we could manually identify.

\subsection{Stylistic sycophancy autorater prompt}
\label{sec:sycophancy-prompt}
\begin{promptbox}
Your task is to determine whether either speaker in the following dialogue exhibits sycophancy.

#Context
        
The dialogue is a multi-turn conversation between two players who each see a different chess board from the same game. Their shared goal is to determine which board came earlier in the game. Because of strict word limits, speakers must summarize selectively and objectively.

# Definition of Sycophancy
        
Sycophancy occurs when a speaker:
1. Uncritically agrees with the other speaker's claims.
2. Repeats the other speaker's statements without adding new information.
3. Validates incorrect or unsubstantiated assertions.
4. Adapts their descriptions to align with the interlocutor despite conflicting evidence.     Sycophancy is about *agreement behavior*, not politeness or cooperation.

# Instructions

Carefully read the full conversation. Focus on whether Alice or Bob:

1. echoes the other's descriptions without independent verification
2. adopts the other's implied conclusion without introducing new evidence
3. validates a factual error or weak inference
4. Do NOT label coordination, neutral restatement, or consistent factual agreement as sycophancy.

#Output Format (STRICT)

Return exactly two lines:

LABEL: <0 | 1 | 2>
JUSTIFICATION: <one sentence>

Where:
0 = No sycophancy detected
1 = Uncritical agreement
2 = Validation of a false or unsupported premise

Do not include any other text.

# Dialogue

    \end{promptbox}
        
\subsection{Information density}
\label{sec:lexical-density-methods}
Let $U=\{u_1, \dots u_n\}$ denote a dialogue containing $u_n$ turns. Let $W$ denote all word tokens in a dialogue and $C\in W$ the subset of context-word tokens. The content ratio captures how much of the text carries semantic content as opposed to disfluencies or function words. We define it as
\begin{equation}
\text{ContentRatio}(U)=\frac{|C|}{|W|}
\end{equation}

 TF–IDF is a standard way to quantify the informativeness of words in a corpus, originally introduced as a measure of term specificity in information retrieval \citep{sparckjones1972idf}. 
 In dialogue, TF-IDF-weighted representations at the utterance level have been widely used for tasks such as dialogue-act classification \citep{Ahmadvand2019ContextualDA}, topic segmentation \citep{matsumoto2022topicbreak}, response evaluation \citep{lowe-etal-2017-towards}, and utterance-level summarization \citep{zhu2006utterance}
 Our novelty score uses TF–IDF-based sentence vectors to identify sentences that introduce new information relative to prior context and therefore rewards dialogues that are both content-heavy and avoid repetition.
 
We treat each utterance $u_i$
 as a separate document in a corpus of size~$n$. Let $\mathbf{v}_i \in \mathbb{R}^d$ denote the TF--IDF vector for $u_i$, with components
 \begin{equation}
\label{eq:tfidf}
v_{ij} = \mathrm{tf}_{ij} \cdot \log\!\left(\frac{n}{\mathrm{df}_{j}}\right),
\end{equation}
where $\mathrm{tf}_{ij}$ is the term frequency of term $j$ in $u_i$ and $\mathrm{df}_{j}$ is the number of utterances in which term $j$ appears. The novelty of an utterance is defined as the mean TF--IDF weight of its non-zero components:
\begin{equation}
\label{eq:novelty}
\mathrm{Novelty}(u_i) =
\begin{cases}
\displaystyle
\frac{1}{|\{j : v_{ij} > 0\}|}
\sum_{j : v_{ij} > 0} v_{ij},
& \text{if } \exists j : v_{ij} > 0, \\[2ex]
0, & \text{otherwise}.
\end{cases}
\end{equation}
And dialogue-level novelty is computed by averaging over utterances:
\begin{equation}
\label{eq:novelty-dialogue}
\mathrm{Novelty}(U)
= \frac{1}{n} \sum_{i=1}^{n} \mathrm{Novelty}(u_i).
\end{equation}
Finally, the lexical density of the dialogue is defined as
\begin{equation}
\label{eq:info-density-details}
\mathrm{LexicalDensity}(U)
= 100 \times
\mathrm{ContentRatio}(U)
\times
\mathrm{Novelty}(U).
\end{equation}

Note that IDF weights are computed globally over all utterances in a dialogue. This means task-relevant terms that recur coherently across turns,  as would occur in a well-grounded dialogue, receive lower IDF weights than terms appearing in only one or two turns. As a side effect, models that terminate early and models that shift topics frequently obtain higher novelty scores than models that develop a coherent line of reasoning across many turns. Common function words and backchannels are excluded prior to computation via the stopword list (see below), so acknowledgement tokens do not directly inflate or deflate scores. 

\paragraph{Non-content words} We computed lexical density using the following list of non-content words, all lower-cased: 

\begin{promptbox}
d, ll, m, re, s, ve, a, about, above, across, after, afterwards, again, against, all, almost, alone, along, already, also, although, always, am, among, amongst, amount, an, and, another, any, anyhow, anyone, anything, anyway, anywhere, are, around, as, at, back, be, became, because, become, becomes, becoming, been, before, beforehand, behind, being, below, beside, besides, between, beyond, both, bottom, but, by, ca, call, can, cannot, could, did, do, does, doing, done, down, due, during, each, eight, either, eleven, else, elsewhere, empty, enough, even, ever, every, everyone, everything, everywhere, except, few, fifteen, fifty, first, five, for, former, formerly, forty, four, from, front, full, further, get, give, go, had, has, have, he, hence, her, here, hereafter, hereby, herein, hereupon, hers, herself, him, himself, his, how, however, hundred, i, if, in, indeed, into, is, it, its, itself, just, keep, last, latter, latterly, least, less, made, make, many, may, me, meanwhile, might, mine, more, moreover, most, mostly, move, much, must, my, myself, n't, name, namely, neither, never, nevertheless, next, nine, no, nobody, none, noone, nor, not, nothing, now, nowhere, n`t, n't, of, off, often, on, once, one, only, onto, or, other, others, otherwise, our, ours, ourselves, out, over, own, part, per, perhaps, please, put, quite, rather, 're, really, regarding, same, say, see, seem, seemed, seeming, seems, serious, several, she, should, show, side, since, six, sixty, so, some, somehow, someone, something, sometime, sometimes, somewhere, still, such, take, ten, than, that, the, their, them, themselves, then, thence, there, thereafter, thereby, therefore, therein, thereupon, these, they, third, this, those, though, three, through, throughout, thru, thus, to, together, too, top, toward, towards, twelve, twenty, two, under, unless, until, up, upon, us, used, using, various, very, via, was, we, well, were, what, whatever, when, whence, whenever, where, whereafter, whereas, whereby, wherein, whereupon, wherever, whether, which, while, whither, who, whoever, whole, whom, whose, why, will, with, within, without, would, yet, you, your, yours, yourself, yourselves, `d, `ll, `m, `re, `s, `ve, 'd, 'll, 'm, 're, 's, 've, mhm, huh, lot, gosh, yep, hm, na, wow, yup, yeah, lots, lot, yes, no, okay, ok, um, uh, hmm, ah, oh, ooh, cuz, uhm, mm, sorry, thanks, aha, ouch, hi, hello, like, just, really, actually, basically, literally, maybe, bit, god, cool,kind, sort, thing, things, stuff, know, mean, think, guess, say, said,  gonna, wanna, gotta, got, get, right, well, anyway, anyways, alright
\end{promptbox}

\subsection{Goal directedness and coherence}
\label{sec:coherence-details}

We implemented a centering theory analysis following the framework of \citet{grosz_joshi_weinstein_1995}. For each dialogue, we extract noun phrases and classify centering transitions between adjacent turns as follows.

\paragraph{Preprocessing} Raw dialogue text is first cleaned to remove formatting artifacts (e.g., literal \texttt{\textbackslash n} sequences and turn numbering). We then parse each utterance using spaCy's \texttt{en\_core\_web\_sm} model.\footnote{\url{https://spacy.io/}} For each sentence within a turn, we extract noun chunks and classify them by grammatical role based on dependency relations:
\begin{itemize}[nosep]
    \item \emph{Subjects}: noun phrases containing a token with dependency label \texttt{nsubj} or \texttt{nsubjpass}
    \item \emph{Objects}: noun phrases containing a token with dependency label \texttt{dobj}, \texttt{iobj}, or \texttt{pobj}
    \item \emph{Neither}: noun phrases that do not fall into either category
\end{itemize}
All noun phrases are normalized to lowercase with punctuation and excess whitespace removed.

\paragraph{Forward-looking centers ($C_f$)} The forward-looking center list for each turn is constructed by concatenating subjects, objects, and other noun phrases in that order, reflecting the standard centering preference for grammatical role salience.

\paragraph{Preferred center ($C_p$)} The preferred center is defined as the first subject if any subjects are present; otherwise, the first object. If neither exists, $C_p$ is undefined.

\paragraph{Backward-looking center ($C_b$)} The backward-looking center links the current turn to the previous turn. We compute $C_b$ by iterating through the previous turn's $C_f$ list in rank order and returning the first entity that also appears in the current turn's $C_f$ list. If no entity is shared, $C_b$ is undefined.

\paragraph{Transition classification} Given the previous turn's backward-looking center ($C_{b_{n-1}}$), the current turn's backward-looking center ($C_{b_n}$), and the current turn's preferred center ($C_{p_n}$), we classify transitions according to Table~\ref{tab:centering_transitions}.

\begin{table}[h]
\small
\begin{tabular}{ll}
\toprule
\textbf{Transition} & \textbf{Condition} \\
\midrule
Continue & $C_{b_n} = C_{b_{n-1}}$ and $C_{b_n} = C_{p_n}$ \\
Retain & $C_{b_n} = C_{b_{n-1}}$ and $C_{b_n} \neq C_{p_n}$ \\
Smooth Shift & $C_{b_n} \neq C_{b_{n-1}}$ and $C_{b_n} = C_{p_n}$ \\
Rough Shift & $C_{b_n} \neq C_{b_{n-1}}$ and $C_{b_n} \neq C_{p_n}$, or $C_{b_n}$ is undefined \\
\bottomrule
\end{tabular}
\caption{Centering transition classification rules.}
\label{tab:centering_transitions}
\end{table}

\paragraph{Centering Coherence Score} We aggregate transition counts into the scalar coherence score defined in Equation~(\ref{eq:cs}), which weights transitions according to the canonical centering preference ordering: Continue (3) $>$ Retain, Smooth Shift (2) $>$ Rough Shift (1).

\paragraph{Algorithm} The complete procedure is summarized below.

\begin{algorithmic}[1]
\Require Dialogue $D = \{u_1, \ldots, u_n\}$
\State $C_b^{\text{prev}} \gets \textsc{None}$; $C_f^{\text{prev}} \gets [\,]$
\For{$i = 1$ \textbf{to} $n$}
    \State Parse $u_i$; extract noun phrases by role: $S_i$ (subjects), $O_i$ (objects), $N_i$ (other)
    \State $C_f^i \gets S_i \oplus O_i \oplus N_i$ \Comment{Concatenate in salience order}
    \State $C_p^i \gets \textsc{First}(S_i)$ if $S_i \neq \emptyset$ else $\textsc{First}(O_i)$
    \State $C_b^i \gets \textsc{First}(\{e \in C_f^{\text{prev}} : e \in C_f^i\})$ \Comment{Highest-ranked shared entity}
    \If{$i = 1$}
        \State $T_i \gets \textsc{Start}$
    \ElsIf{$C_b^i = \textsc{None}$}
        \State $T_i \gets \textsc{Rough Shift}$
    \ElsIf{$C_b^i = C_b^{\text{prev}}$}
        \State $T_i \gets \textsc{Continue}$ if $C_b^i = C_p^i$ else $\textsc{Retain}$
    \Else
        \State $T_i \gets \textsc{Smooth Shift}$ if $C_b^i = C_p^i$ else $\textsc{Rough Shift}$
    \EndIf
    \State $C_f^{\text{prev}} \gets C_f^i$; $C_b^{\text{prev}} \gets C_b^i$
\EndFor
\State \Return Transition sequence $\{T_1, \ldots, T_n\}$
\end{algorithmic}

\end{document}